\documentclass[10pt,twocolumn,letterpaper]{article}

\usepackage{cvpr}              

\usepackage{graphicx}
\usepackage{amsmath}
\usepackage{amssymb}
\usepackage{booktabs}

\usepackage{url}
\usepackage{tabularx}
\usepackage{comment}
\usepackage{makecell}
\usepackage{pifont}
\usepackage{multirow}
\usepackage[misc]{ifsym}
\usepackage{afterpage}
\usepackage{setspace}
\usepackage{breakcites}
\usepackage{seqsplit}
\usepackage{times}

\newcommand{\myparagraph}[1]{\vspace{0.1em}\noindent\textbf{#1}}

\newcommand{\fk}[1]{\textcolor{black}{#1}}

\usepackage[pagebackref=true,breaklinks=true,colorlinks,bookmarks=false]{hyperref}

\usepackage[capitalize]{cleveref}
\crefname{section}{Sec.}{Secs.}
\Crefname{section}{Section}{Sections}
\Crefname{table}{Table}{Tables}
\crefname{table}{Tab.}{Tabs.}

\begin{document}

\title{A Large-Scale Outdoor Multi-modal Dataset and Benchmark
for \\Novel View Synthesis and Implicit Scene Reconstruction}


\author{
    Chongshan Lu$^{1}$ \quad\quad
    Fukun Yin$^{1,2}$ \quad\quad
    Xin Chen$^{2}$ \quad\quad
    Tao Chen$^{1}$\thanks{Corresponding author.} \quad\quad
    Gang Yu$^{2}$ \quad\quad
    Jiayuan Fan$^{1}$
    \\
    $^{1}$Fudan University \quad
    $^{2}$Tencent PCG \quad
    \\
    \tt \small \textbf{\href{https://ommo.luchongshan.com}{https://ommo.luchongshan.com}}
}

\maketitle

\begin{abstract}

Neural Radiance Fields (NeRF) has achieved impressive results in single object scene reconstruction and novel view synthesis, as demonstrated on many single modality and single object focused indoor scene datasets like DTU~\cite{jensen2014dtu}, BMVS~\cite{yao2020blendedmvs}, and NeRF Synthetic~\cite{mildenhall2020nerf}. 
However, the study of NeRF on large-scale outdoor scene reconstruction is still limited, as there is no unified outdoor scene dataset for large-scale NeRF evaluation due to expensive data acquisition and calibration costs. 

In this work, we propose a large-scale \textbf{o}utdoor \textbf{m}ulti-\textbf{mo}dal dataset,\textbf{ OMMO dataset}, containing complex objects and scenes with calibrated images, point clouds and prompt annotations. A new benchmark for several outdoor NeRF-based tasks is established, such as novel view synthesis, surface reconstruction, and multi-modal NeRF. To create the dataset, we capture and collect a large number of real fly-view videos and select high-quality and high-resolution clips from them. Then we design a quality review module to refine images, remove low-quality frames and fail-to-calibrate scenes through a learning-based automatic evaluation plus manual review. Finally, a number of volunteers are employed to add the text descriptions for each scene and keyframe.
Compared with existing NeRF datasets, our dataset contains abundant real-world urban and natural scenes with various scales, camera trajectories, and lighting conditions. Experiments show that our dataset can benchmark most state-of-the-art NeRF methods on different tasks. We will release the dataset and model weights soon.

\end{abstract}


\section{Introduction}

\begin{figure}[tbp] 
	\centering 
	\includegraphics[width=\linewidth]{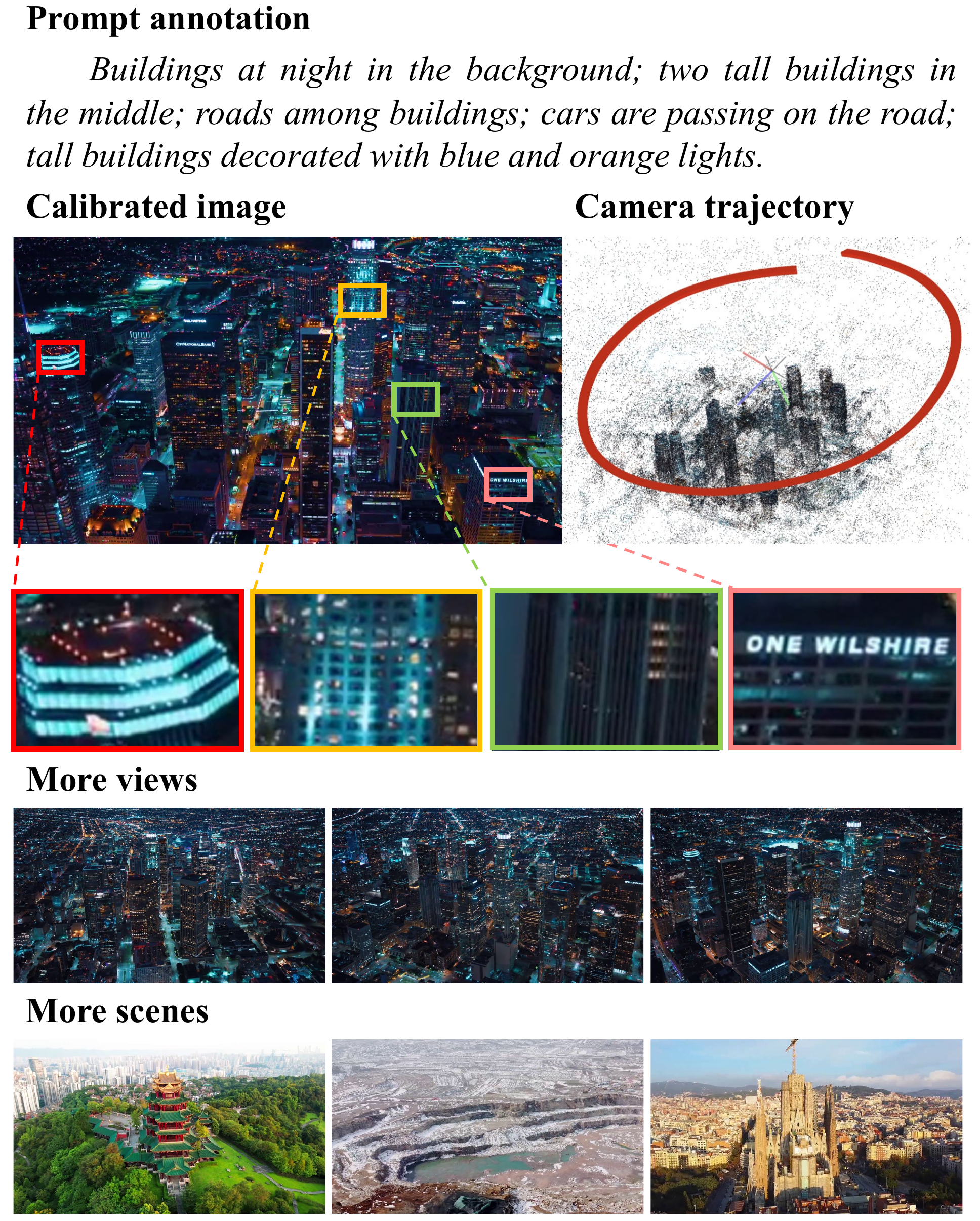} 
	\caption{A city scene example from our dataset captured with low illuminance and circle-shaped camera trajectory. We show multi-view calibrated images, the camera track, and text descriptions of the scene. Some details in colored boxes are zoomed in to indicate that our dataset can provide real-world high-fidelity texture details.} 
	\label{fig:teaser} 
\end{figure} 

Recent advances in implicit neural representations have achieved remarkable results in photo-realistic novel view synthesis and high-fidelity surface reconstruction~\cite{yin2022coordinates,yariv2021volume}.
Unfortunately, most of the existing methods focus on a single object or an indoor scene~\cite{yin2022coordinates,yariv2021volume,deng2022depth,johari2022geonerf,chen2022geoaug}, and their synthesis performance will decrease drastically if migrated to outdoor scenes. Although some very recent methods try to solve this problem and are well-designed for large scenes~\cite{turki2022mega,xiangli2022bungeenerf}, their performance is difficult to compare due to the lack of large-scale outdoor scene datasets and uniform benchmarks.

At present, the existing outdoor scene datasets are either collected with simple scenes containing very few objects, or rendered from virtual scenes, all at a small geographical scale. For example, Tanks and temples~\cite{knapitsch2017tanks} provides a benchmark of realistic outdoor scenes captured by a high-precision industrial laser scanner, but its scene scale is still too small (463$m^{2}$ on average) and only focuses on a single outdoor object or building. The BlendedMVS~\cite{yao2020blendedmvs} and UrbanScene3D~\cite{liu2021urbanscene3d} datasets contain scene images rendered from reconstructed or virtual scenes, which deviate from the real scene in both texture and appearance details. Collecting images from the Internet can theoretically build very effective datasets~\cite{heinly2015reconstructing,agarwal2011building}, like ImageNet~\cite{deng2009imagenet} and COCO~\cite{lin2014microsoft}, but these methods are not suitable for NeRF-based task evaluation due to the changes of objects and lighting conditions in the scene at different times. Our dataset acquisition method is similar to Mega-NeRF~\cite{turki2022mega}, which captures large real-world scenes by drones. But Mega-NeRF only provides two monotonic scenes, which hinders it from being a widely used baseline. 
Therefore, to our knowledge, no uniform and widely recognized large-scale scene dataset is built for NeRF benchmarking, causing large-scale NeRF research for outdoor far fall behind that for single objects or indoor scenes~\cite{jensen2014dtu,yao2020blendedmvs,mildenhall2020nerf,dai2017scannet}.

\begin{table}[t]
	\begin{center}
		\centering
		\caption{Comparison with existing NeRF datasets, especially those outdoor datasets (or outdoor parts) related to ours. The first group is for single objects, the second group is for large scenes, and the last row is our dataset. For each dataset, we show the number of scenes and images, \fk{whether the scene types, camera trajectories, and lighting conditions are diverse,} whether they are real-world scenes (called Real), and whether they have multi-modal data (called M-modal).}
		\label{tab:Comparisonexistingdatasets}
		\resizebox{0.5\textwidth}{!}{
		\begin{tabular}{lccccccc}
			\toprule
			Datasets      & \# Scenes &  \# Images  & Types &  Camera & Lighting & Real &M-modal  \\
			\midrule
			DTU~\cite{jensen2014dtu} & 124 & 4.2K  & No & No & Yes & Yes & No \\
			NeRF~\cite{mildenhall2020nerf} &18  &3551  &No  &Yes &No &Yes &No  \\
			Scannet~\cite{dai2017scannet} & 1.5K &  2.5M  & No   & Yes & No & Yes  & No\\
			\hline
			T \& T~\cite{knapitsch2017tanks}         & 6  & 88k   &No &No &No & Yes    &No  \\
			
			BlendedMVS ~\cite{yao2020blendedmvs}        & 28 & 5k   & Yes   & No & No & No & No \\
			UrbanScene3D~\cite{liu2021urbanscene3d} &  16 & 10.4K  & Yes & No & No & Part & No \\
			Quad 6k~\cite{crandall2011discrete} & 1 & 5.1K & No & Yes & No & Yes & No  \\
			Mill 19~\cite{turki2022mega} & 2 & 3.6K   & Yes & No &No&Yes&No \\
			
			\midrule
			\textbf{Ours} & 33 & 14.7K & Yes & Yes & Yes & Yes & Yes \\
			\bottomrule
		\end{tabular}
		}
	\end{center}
\end{table}

To address the lack of large-scale real-world outdoor scene datasets, we introduce a well-selected fly-view multi-modal dataset. The dataset contains totally 33 scenes with prompt annotations, tags, and 14K calibrated images, as shown in Figure \ref{fig:teaser}. 
Different from the existing methods mentioned above, the sources of our scenes are very extensive, including those collected on the Internet and  captured by ourselves. Meanwhile, the collection indicators are also comprehensive and representative, including various scene types, scene scales, camera trajectories, lighting conditions, and multi-modal data that are not available in existing datasets (see Table ~\ref{tab:Comparisonexistingdatasets}. More importantly, we provide a generic pipeline to generate real-world NeRF-based data from drone videos on the Internet, which makes our dataset easily to be extensible by the community. 

Further, to evaluate the applicability and performance of the built dataset for evaluating mainstream NeRF methods, we build all-around benchmarks including novel view synthesis, scene representations, and multi-modal synthesis based on the dataset. Moreover, \fk{we provide several detailed sub-benchmarks for each above task, according to different scene types, scene scales, camera trajectories and lighting conditions, to give a fine-grained evaluation of each method.}

\begin{figure}[t]
	\centering
	\includegraphics[width=0.96\linewidth]{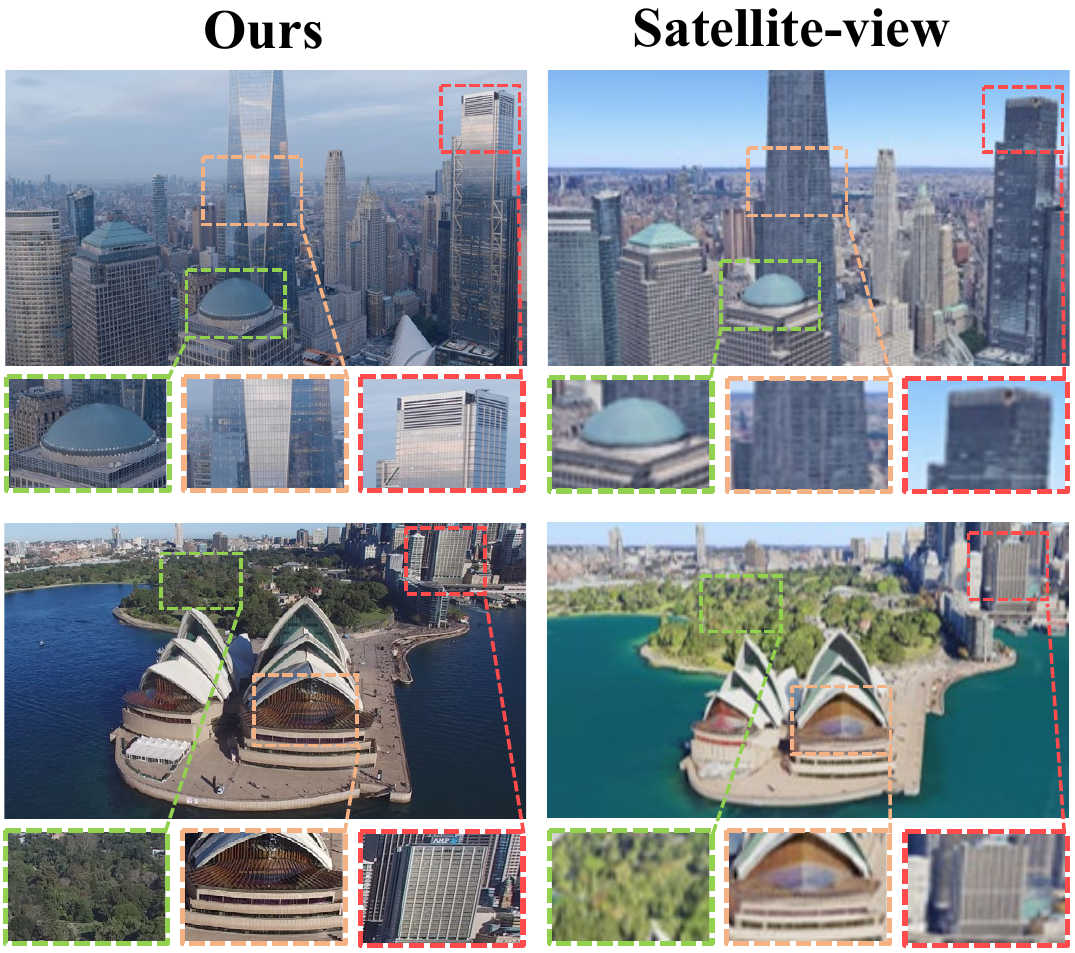}
	\caption{Visual comparison with existing large-scale satellite-view outdoor datasets~\cite{xiangli2022bungeenerf} acquired from Google Earth Studio. The top row is from~\cite{xiangli2022bungeenerf}, and the bottom row is corresponding scenes from our fly-view dataset, which is more realistic with clear textures and rich details (zoom-in for the best of views).}
	\label{fig:division}
\end{figure}

\begin{figure*}[t]
	\centering
	\includegraphics[width=\linewidth]{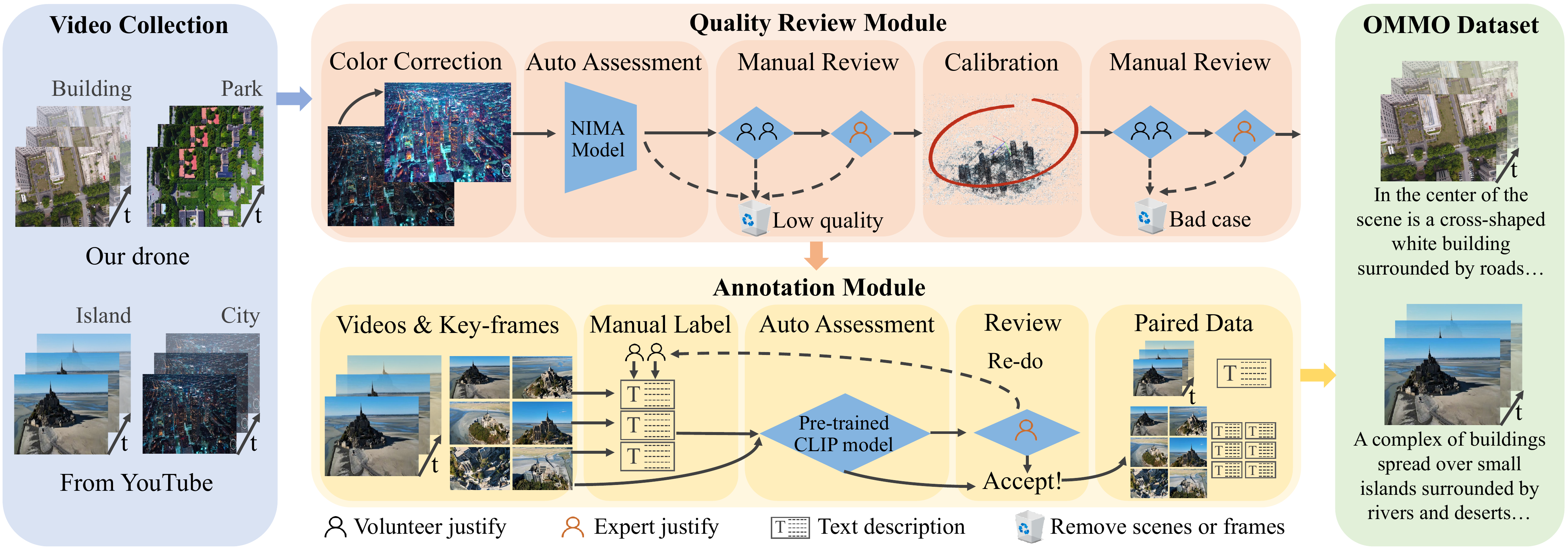}
	\caption{The pipeline for our dataset generation. The original videos are collected from both YouTube and captured by us, and then fed into the review and annotation module. The former mainly removes low-quality frames and failed scenes; the latter annotates text descriptions for scenes and keyframes. Currently, we have generated 33 scenes with 14K images and text descriptions.}
	\label{fig:pipeline}
\end{figure*}

To summarize, our main contributions include: 
\begin{itemize} 
\setlength\itemsep{0em}
    \item Aiming at advancing the large-scale NeRF research, we introduce an outdoor scene dataset captured from the real world with multi-modal data, which surpasses all existing relative outdoor datasets in both quantity and diversity, see Table \ref{tab:Comparisonexistingdatasets} and Sec.~\ref{paper:divisionmethod}.
	
    \item To form a uniform benchmarking standard for outdoor NeRF methods, we create multiple benchmark tasks for mainstream outdoor NeRF methods. Extensive experiments show that our dataset can well support common NeRF-based tasks and provide prompt annotations for future research, see Sec.~\ref{ExperimentalResults}.

    \item We provide a cost-effective pipeline for converting videos that can be flexibly accessed from the Internet to NeRF-purpose training data, \fk{which makes our dataset easily scalable}, see Sec.~\ref{paper:acquisitionmethod} and Sec.~\ref{paper:calibrationmethod}.
    
\end{itemize}


\section{Related Work}

\subsection{Neural Scene Representation and Rendering}

Neural Radiance Fields (NeRF) ~\cite{mildenhall2020nerf} propose an effective implicit neural scene representation method to synthesize novel views by the single scene optimization. Many subsequent coordinate-based methods are inspired by it, and we can classify them into neural surface fields~\cite{oechsle2021unisurf,yariv2021volume,wang2021neus} and neural radiance fields~\cite{niemeyer2022regnerf, kania2022conerf,zhang2020nerf++} based on the shape representation differences. However, both ways have poor performance on large scenes due to the limited representation capability of MLP-based networks adopted in NeRF. Even with prior information, it is difficult to directly apply small-scale scene-focused methods due to the increased scene complexity~\cite{yin2022coordinates,kulhanek2022viewformer}.

Fortunately, some very recent works have started to study the neural representation for large-scale scenes. Mega-NeRF~\cite{turki2022mega} divides the large fly-view scene into multiple small blocks to train specialized NeRFs in parallel. Block-NeRF~\cite{tancik2022block} also adopts this simplified idea, dividing the neighborhood into blocks, and then novel views are sampled from overlapping blocks and combined according to inverse distance weights. CityNerf (BungeeNeRF)~\cite{xiangli2022bungeenerf} introduces a progressive neural radiance field that starts from fitting distant views with a shallow base block, and appends new blocks to accommodate details. NeRF in the Wild (NeRF-W) ~\cite{martin2021wild} introduces a series of extensions to NeRF~\cite{mildenhall2020nerf} to synthesize novel views of complex scenes, using only unstructured collections of in-the-wild photos. Recursive-NeRF~\cite{yang2022recursive} provides an efficient and adaptive rendering and training approach for NeRF, that forwards high uncertainties coordinates to a bigger network with more powerful representational capability. 

However, the aforementioned large-scale NeRF methods~\cite{turki2022mega,tancik2022block,xiangli2022bungeenerf,martin2021wild,yang2022recursive}use different outdoor datasets with various capturing conditions and research focuses, causing difficulty to fairly compare these NeRF methods' performance on common tasks based on a uniform benchmark.

\subsection{NeRF-based Datasets and Benchmarks}

There are widely used NeRF datasets and established benchmarks for single objects~\cite{jensen2014dtu,mildenhall2020nerf}, unbounded objects~\cite{yao2020blendedmvs}, human faces~\cite{ramon2021h3d}, and indoor scenes~\cite{dai2017scannet}, see the first group in Table \ref{tab:Comparisonexistingdatasets}. 

For large-scale outdoor scenes, some datasets provide high-fidelity models from accurate radar scans, but this expensive data acquisition makes the scale and size of these datasets still unsatisfactory~\cite{knapitsch2017tanks}. Rendering images from optimized models will result in higher cost and unrealistic scenes~\cite{yao2020blendedmvs,liu2021urbanscene3d}. Correspondingly, a low-cost and easy-to-expand way is to collect images of the same scene shot by different people and devices, at different times from the Internet~\cite{heinly2015reconstructing,agarwal2011building}. However, these methods do not meet the needs of common NeRF tasks due to the changes in weather, lighting, and objects in the scene. Mega-NeRF~\cite{turki2022mega} builds two high-quality fly-view real scenes and calibrated images, but has not become a widely used benchmark due to its small size and single type. There are also large-scale NeRF datasets and benchmarks for specific problems, such as neighborhoods~\cite{tancik2022block} or remote sensing~\cite{xiangli2022bungeenerf}, see the second group in Table \ref{tab:Comparisonexistingdatasets}.

In conclusion, for the reasons mentioned above, none of the above datasets have formed a widely used uniform benchmark. So a more comprehensive outdoor NeRF-based dataset is required, to facilitate the research and exploitation of larger-scale implicit scene representation. In contrast, the built dataset in this work provides 33 large-scale scenes, more than 14K fly-view images with camera poses, rich content, and text descriptions. Meanwhile, several new large-scene fly-view-based benchmarks for novel view synthesis, implicit scene representations, and multi-modal synthesis tasks are also proposed.

\section{Dataset Generation}
\label{paper:datasetgeneration}

Our dataset acquisition, calibration, and annotation pipeline are shown in Figure \ref{fig:pipeline}. We first decompose and enhance the original videos by time-sampling and color-correcting, 
and review the quality of the frames 
and scene calibrations by automated models followed by volunteers to remove low-quality frames and fail-to-calibrate scenes, see Sec.~\ref{paper:acquisitionmethod}. Then volunteers provide prompt annotations for each scene and keyframe, and the CLIP~\cite{radford2021learning} model is exploited to cooperate with human experts, to supervise the semantic consistency between labeled text and corresponding images, see Sec.~\ref{paper:calibrationmethod}. Finally, we introduce the dataset distribution from several aspects such as scene categories and collection cost, see Sec.\ref{paper:divisionmethod} and Sec.\ref{paper:cost}.

\begin{figure*}[t]
	\centering
	\includegraphics[width=1\linewidth]{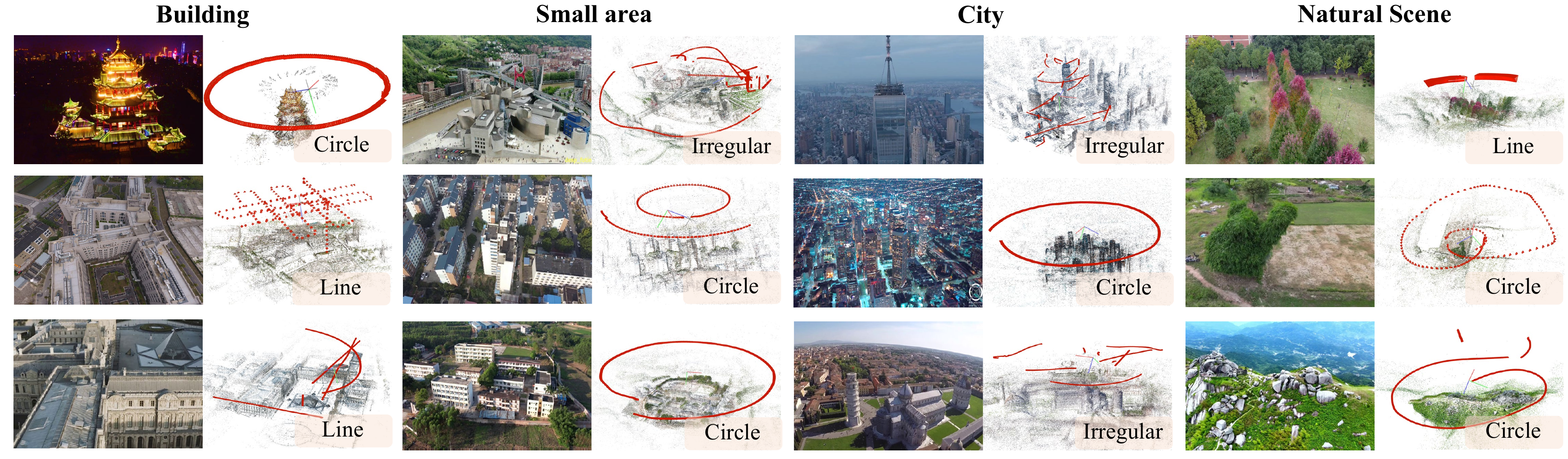}
	\caption{Examples of different types from our dataset. We visualize some scenes and camera trajectories from our dataset, which contain both urban and natural scenes with various scales, camera trajectories, and lighting conditions.}
	\label{fig:division}
\end{figure*}

\subsection{Acquisition and Calibration Method}
\label{paper:acquisitionmethod}
\textbf{Original Videos.} Our outdoor fly-view dataset mainly comes from two sources: captured by ourselves and collected on YouTube. Among them, the YouTube videos are from worldwide, including urban and natural scenes of various geographical scales, various camera moving trajectories and lighting conditions. However, these videos often have limited resolution due to the compression when uploading, while our captured videos are all 4K HD and can meet more high-fidelity NeRF needs. Benefiting from the diversity of the videos, the proposed scenes in our dataset include various elements, such as buildings, roads, trees, islands, mountains, rivers, etc. In addition, compared to synthetic data, our scenes' space layout, surface reflection, and lighting conditions are completely real, which supports NeRF methods' learning about real-world outdoor scenes. To have different lighting conditions in the same scene like DTU~\cite{jensen2014dtu}, we use drones to capture videos of the same building at different times and weather conditions. To explore the impact of different camera trajectories, we scan the same scene with different flight strategies. Overall, our original videos come from 268 real scans from around the world, covering a variety of scene types, camera trajectories, and lighting conditions.

\textbf{Color Correction.} We set different sampling intervals according to the video length and frames per second, so that the remaining frame number is between 800-1000 to meet the needs of calibration, yielding a total of 240K frames. For low-light or rainy scenes, we first use an image enhancement model to recover the texture features~\cite{kim2021representative}.

\textbf{Auto Assessment.} Then NIMA~\cite{talebi2018nima} model is employed to evaluate image quality and remove blur, ghosting, and low-quality images. After this step, the 240K frames from 268 scenes in the above step are left with 152K frames.

\textbf{Manual Quality Review.} After the quality auto-evaluation, there are still some frames with low quality \fk{i.e. blurring, artifacts, or focusing outside the scene}. So we employ three volunteers to manually discard frames that do not meet requirements. Specifically, the three volunteers consist of 2 professional data labelers and a domain expert. For each frame, 2 professional data labelers judge whether to keep it or not, and the decision is made if both labelers agree, otherwise it is up to the domain expert for the decision. After this step, there are still 110K frames left.

\textbf{Calibration.} Large-scale scene calibration based on highly dynamic images has always been a difficult problem, especially for outdoor scenes with no obvious local detail differences~\cite{bozic2021transformerfusion,legros2020robust,guo2022neural}. As a common solution, we use COLMAP~\cite{schonberger2016structure,schonberger2016pixelwise} to achieve multi-view 3D reconstruction, image calibration and depth image rendering. 
\fk{It is foreseeable that the reconstruction of some scenes with insufficient overlap and textures, or forwardly moving camera motion will fail. These fail-to-calibrate scenes cannot meet the requirements of NeRF-based methods, which need to be removed manually.}

\textbf{Manual Scene Review.} We invite the above three volunteers to review the calibration quality based on the completeness of scene point clouds, following the same decision-making process. A large number of scenes are discarded at this step, and finally, 33 real-world scans with nearly 14K calibrated images constitute our outdoor dataset, which surpasses existing large-scale datasets in both quantity and diversity.

\subsection{Prompt Annotation Method}
\label{paper:calibrationmethod}

To provide prompt annotations for multi-modal NeRF, we add text descriptions to each scene and keyframes. Note that generating descriptions is a more subjective and time-consuming task, so we employ more volunteers and apply pre-trained CLIP~\cite{radford2021learning} models.

\textbf{Manual Label.} Specifically, six trained volunteers participate in this progress, who non-repeatedly extract and record the corresponding descriptive texts from scenes and keyframes, respectively. These volunteers include high-year Ph.D. students and professors in computer vision and natural language processing fields who can handle this job well.

\textbf{Auto Assessment.} Each frame and annotated text is fed into a CLIP~\cite{radford2021learning} model pre-trained on large scene-text datasets~\cite{sanabria2018how2,grubinger2006iapr} to compute their similarity scores. If the similarity score is above the threshold, we accept this annotation. Otherwise, we leave it to experts to double-check.

\textbf{Expert Review.} For annotations that the CLIP model cannot judge, experts will evaluate whether the text can describe the image comprehensively and objectively. If possible, we accept the label, otherwise, we hand it over to another volunteer to label again. Fortunately, the frames and scenes in our dataset are only re-annotated at most 2 times.

We have annotated 33 scans with corresponding descriptions and tags, and part of keyframes, which can well meet the training needs of multi-modal NeRF. We are still annotating the remaining keyframes for more potentially complex needs.

\subsection{Distribution}
\label{paper:divisionmethod}

\begin{table}[t]
	\begin{center}
		\centering
		\caption{Distribution of our dataset. We divide our dataset into subsets based on scene type, camera trajectory, and lighting condition and count the number of each subset.}
		\label{tab:divisionmethod}
		\resizebox{0.48\textwidth}{!}{
		\begin{tabular}{@{}cccccc@{}}
			\toprule
			\thead{Scene\\Type}      & \# Scenes &  \thead{Camera \\Trajectory}  &\# Scenes & \thead{Lighting \\Condition} & \# Scenes  \\
                \cmidrule(l){1-2} \cmidrule(l){3-4} \cmidrule(l){5-6} 					
			Building &  8& Circle   & 15   & Day&30 \\
			Small area & 9 & Line   & 10   & Night&3\\
			City & 8   & Irregular   & 8& \\
			Natural scene         & 8  &    &    &  \\
			\bottomrule
		\end{tabular}
		}
	\end{center}
\end{table}

According to different division methods, the distribution of our dataset is shown in Table ~\ref{tab:divisionmethod}. However, some scene types are relatively ambiguous. For example, when a building is surrounded by plenty of trees, warehouses, etc., it is difficult to say whether such an image belongs to a building type or not. To resolve the ambiguity, we design a questionnaire and invite 50 voters to determine the attributes of the scenes. Invited volunteers range from 19 to 53 years old, and we recommend a very typical reference for each attribute.

Our dataset contains both natural and urban scenes, which are further divided into buildings, small areas, and whole cities. The performance on different subsets can verify the most suitable scene type and scale for each NeRF method. Some methods only target scenes with camera trajectories moving in a ring or matrix. For a fair baseline, we divide the scene into circles, lines, and irregular, according to the camera trajectories. In addition, few recent methods focus on low-illuminance NeRF research, so we also provide scenes with different lighting conditions. In particular, we collect some scans of the same scene under different lighting conditions, which can evaluate the ability of the method against poor lighting.

We provide benchmarks for novel view synthesis, generalization, scene representations, and multi-modal synthesis tasks. At the same time, according to the above scene types, our dataset can also produce  corresponding sub-benchmarks to evaluate methods under different conditions and settings, see Sec.~\ref{ExperimentalResults}.

\begin{figure*}[t]
	\centering
	\includegraphics[width=1\linewidth]{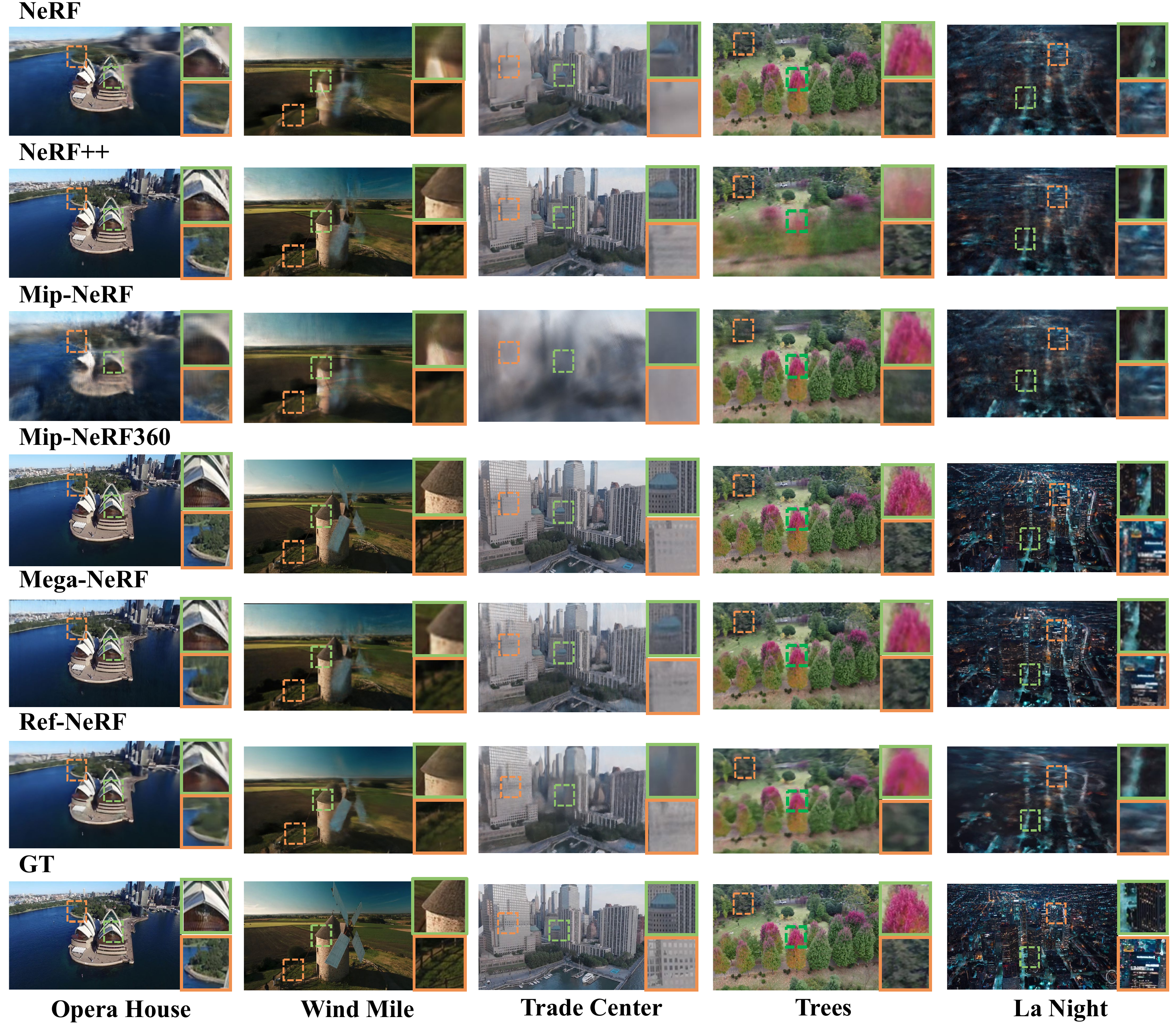}
	\vspace{-15pt}
	\caption{Qualitative visualization results for novel view synthesis (zoom-in for the best of views) on our OMMO dataset.}
	\label{fig:exp}
\end{figure*}

\subsection{Cost}
\label{paper:cost}

Since our dataset requires volunteers to review images and scenes' quality and annotate the scenes and keyframes with texts, it is unavoidably time-consuming and labor-intensive.

Data collection which involves drone purchasing and shooting, and computing introduced by pre-training scene-text CLIP model, account for a large part of the total cost. But once the drone and CLIP model is ready, we can just use them to prepare more new scenes and add to the dataset, without involving additional cost.

More importantly, with the increase of labeled data that can help train a better model, we can replace part of volunteer review and annotation work with state-of-the-art trained models to optimize our pipeline. Therefore, although our method may be expensive in the early stage, it has good potential to be automated and save future costs that may be introduced when expanding the dataset.

\begin{table*}[t]
	\begin{center}
		\centering
		\caption{Benchmark for novel view synthesis. We present the performance of six state-of-the-art and representative methods on our dataset. $\uparrow$ means the higher, the better.}
		\label{tab:novelviewsynthesis}
		\resizebox{1\textwidth}{!}{
		\begin{tabular}{c|ccc|ccc|ccc|ccc|ccc|ccc|ccc}
			\toprule
			\multirow{2}{*}{Scene ID}  & Scene & Camera  & Lighting & \multicolumn{3}{c|}{\textbf{NeRF}~\cite{mildenhall2020nerf}}& 
			\multicolumn{3}{c|}{\textbf{NeRF++}~\cite{zhang2020nerf++}}& 
			\multicolumn{3}{c|}{\textbf{Mip-NeRF}~\cite{barron2021mip}}&
			\multicolumn{3}{c|}{\textbf{Mip-NeRF 360}~\cite{barron2022mip}}&
			\multicolumn{3}{c|}{\textbf{Mega-NeRF}~\cite{turki2022mega}}& 
			\multicolumn{3}{c}{\textbf{Ref-NeRF}~\cite{xiangli2022bungeenerf}}\\
			
			 & Types & Tracks &Conditions & PSNR$\uparrow$&SSIM$\uparrow$&LPIPS$\downarrow$& PSNR$\uparrow$&SSIM$\uparrow$&LPIPS$\downarrow$& PSNR$\uparrow$&SSIM$\uparrow$&LPIPS$\downarrow$&
			 PSNR$\uparrow$&SSIM$\uparrow$&LPIPS$\downarrow$&
			 PSNR$\uparrow$&SSIM$\uparrow$&LPIPS$\downarrow$& PSNR$\uparrow$&SSIM$\uparrow$&LPIPS$\downarrow$\\

            \midrule

			1 & Buildings & Irregular & Day 
			&16.93&0.369&0.744
			&16.86&0.359&0.780
			&16.84&0.369&0.793
			&13.91&0.311&0.771
			&16.12&0.341&0.782
			&15.10&0.344&0.755\\
            2 & Small area & Circles & Day 
            &15.31&0.442&0.694 
            &14.89&0.471&0.653
            &15.16&0.396&0.731
            &15.06&0.438&0.646
            &15.64&0.467&0.679
            &15.90&0.490&0.632
            \\
            3 & Citys & Lines & Day 
            &14.38 &0.278&0.556 
            &14.64&0.294&0.547
            &14.56&0.288&0.533
            &14.25&0.309&0.526
            &15.21&0.325&0.517
            &15.44&0.371&0.526\\
            4 & Buildings & Circle & Night 
            &25.39 &0.859 &0.431 
            &27.47&0.898&0.380
            &21.78&0.758&0.469
            &27.68&0.943&0.292
            &23.36&0.855&0.419
            &27.86&0.905&0.404\\
            5 & Small area & Circles & Day 
            &22.26 &0.670 &0.531 
            &24.32&0.729&0.450
            &14.98&0.544&0.633
            &25.76&0.801&0.317
            &25.78&0.763&0.436
            &23.54&0.706&0.491\\
            6 & Natural scenes & Circles & Day 
            &24.09 &0.679 &0.504 
            &25.59&0.749&0.396
            &23.18&0.658&0.529
            &28.86&0.896&0.211
            &24.92&0.772&0.393
            &26.07&0.716&0.459\\
            7 & Buildings & Lines & Day 
            &5.36 &0.166 &0.747 
            &21.93&0.707&0.542
            &15.57&0.643&0.624
            &23.05&0.734&0.523
            &22.33&0.691&0.552
            &25.79&0.731&0.511\\
            8 & Citys & Circle & Day 
            &21.14 &0.496 &0.594 
            &22.91&0.568&0.509
            &19.82&0.462&0.638
            &25.07&0.714&0.354
            &16.65&0.478&0.431
            &21.21&0.489&0.606\\
            9 & Citys & Lines & Day 
            &14.92&0.344&0.744
            &14.57&0.341&0.732
            &14.58&0.338&0.746
            &15.40&0.303&0.706
            &17.32&0.491&0.673
            &20.34&0.432&0.649
            \\
            10 & Citys & Irregular & Day 
            &22.26 &0.550 &0.626 
            &24.37&0.599&0.578
            &19.80&0.528&0.643
            &26.68&0.719&0.420
            &21.78&0.615&0.558
            &24.23&0.578&0.597\\
            11 & Buildings & Circles & Night 
            &22.36 &0.816 &0.420 
            &24.61&0.852&0.342
            &22.81&0.822&0.423
            &27.06&0.931&0.217
            &24.37&0.844&0.392
            &23.81&0.843&0.355\\
            12 & Small area & Circles & Day 
            &22.41 &0.594 &0.533
            &24.29&0.675&0.447
            &22.13&0.601&0.526
            &28.12&0.825&0.274
            &21.60&0.619&0.493
            &23.06&0.604&0.524\\
            13 & Buildings & Lines & Day 
            &22.27 &0.592 &0.608 
            &23.52&0.623&0.581
            &18.90&0.537&0.673
            &26.63&0.771&0.403
            &25.50&0.722&0.517
            &23.29&0.605&0.594\\
            14 & Small area & Lines & Day 
            &19.85 &0.554 &0.569 
            &23.89&0.737&0.417
            &17.06&0.481&0.655
            &28.06&0.894&0.224
            &24.42&0.746&0.411
            &21.76&0.625&0.508\\
            15 & Small area & Circles & Day 
            &20.35&0.527&0.552 
            &21.71&0.612&0.490
            &19.44&0.489&0.594
            &28.63&0.888&0.179
            &22.69&0.665&0.445
            &20.33&0.497&0.576
            \\
            16 & Natural scenes & Circles & Day 
            &17.86&0.397&0.631
            &18.75&0.405&0.597
            &18.49&0.399&0.610
            &10.01&0.344&0.850
            &20.26&0.532&0.509
            &19.64&0.428&0.572\\
            17 & Natural scenes & Circles & Day 
            &22.02&0.571&0.610
            &24.20&0.671&0.461
            &17.01&0.526&0.696
            &29.53&0.833&0.247
            &17.23&0.574&0.529
            &23.17&0.589&0.529\\
            18 & Small area & Lines & Day 
            &26.06 &0.754 &0.428 
            &25.57&0.730&0.461
            &24.61&0.732&0.469
            &28.55&0.855&0.265
            &24.76&0.733&0.448
            &22.79&0.674&0.569\\
            19 & Small area & Circles & Day 
            &14.20&0.399&0.726
            &13.86&0.373&0.703
            &13.84&0.394&0.738
            &14.72&0.367&0.676
            &23.81&0.682&0.465
            &14.34&0.386&0.691\\
            20 & Citys & Circles & Day 
            &22.84 &0.613 &0.499 
            &23.28&0.642&0.475
            &22.41&0.603&0.519
            &28.33&0.862&0.228
            &21.11&0.633&0.490
            &21.54&0.553&0.574\\
            21 & Natural scenes & Circles & Day 
            &22.59&0.514&0.532
            &21.84&0.473&0.593
            &22.31&0.513&0.537
            &25.64&0.747&0.344
            &21.92&0.506&0.578
            &21.07&0.436&0.672\\
            22 & Buildings & Lines & Day 
            &16.53 &0.466 &0.733
            &20.66&0.558&0.575
            &13.37&0.420&0.776
            &24.79&0.766&0.362
            &20.84&0.597&0.527
            &20.31&0.530&0.615
            \\
            23 & Natural scenes & Lines & Day 
            &18.99&0.405&0.669
            &19.51&0.417&0.597
            &18.09&0.389&0.671
            &21.25&0.514&0.539
            &20.13&0.438&0.585
            &19.94&0.409&0.622\\
            24 & Natural scenes & Lines & Day 
            &19.32&0.386&0.696
            &23.14&0.522&0.535
            &16.89&0.374&0.715
            &25.86&0.707&0.373
            &23.87&0.563&0.518
            &22.17&0.452&0.616\\
            25 & Natural scenes & Lines & Day 
            &24.72&0.550&0.528
            &22.42&0.509&0.613
            &24.24&0.541&0.542
            &28.91&0.789&0.306
            &25.98&0.629&0.457
            &23.62&0.502&0.598
            \\
            26 & Buildings & Irregular & Day 
            &8.56 &0.242 &0.564 
            &19.94&0.586&0.513
            &13.43&0.353&0.688
            &14.59&0.459&0.626
            &19.23&0.669&0.467
            &21.00&0.615&0.489
            \\
            27 & Citys & Irregular & Day 
            &4.54&0.006&0.705
            &21.25&0.548&0.546
            &14.82&0.453&0.674
            &21.26&0.599&0.235
            &20.59&0.606&0.543
            &20.82&0.519&0.590\\
            28 & Small area & Circles & Day 
            &24.48 &0.660 &0.479 
            &23.28&0.642&0.475
            &24.76&0.659&0.406
            &29.62&0.874&0.240
            &25.87&0.723&0.442
            &22.17&0.452&0.616\\
            29 & Buildings & Circle & Day 
            &22.98 &0.608 &0.540 
            &23.17&0.617&0.529
            &23.01&0.609&0.539
            &25.51&0.740&0.400
            &21.57&0.611&0.557
            &21.11&0.543&0.631\\
            30 & Natural scenes & Irregular & Day 
            &20.23 &0.522 &0.605 
            &23.27&0.639&0.476
            &18.63&0.461&0.675
            &26.54&0.837&0.296
            &24.04&0.686&0.459
            &21.62&0.535&0.586
            \\
            31 & Citys & Circles & Night 
            &18.97&0.365&0.645
            &19.05&0.371&0.643
            &18.91&0.358&0.659
            &13.08&0.234&0.708
            &20.93&0.596&0.545
            &19.18&0.372&0.645\\
            32 & Citys & Irregular & Day 
            &17.99&0.582&0.621
            &18.99&0.605&0.540
            &11.28&0.424&0.687
            &17.16&0.566&0.601
            &21.29&0.702&0.475
            &18.98&0.595&0.565\\
            33 & Citys & Irregular & Day 
            &5.79 &0.007 &0.745 
            &20.19&0.497&0.597
            &14.31&0.42&0.755
            &22.76&0.629&0.457
            &22.89&0.635&0.478
            &21.23&0.522&0.578\\

            \midrule
            \textbf{Mean} &-&-&-
            &18.72 &0.484 &0.600
            &21.45&0.576&0.538
            &18.39&0.501&0.623
            &23.10&0.672&0.418
            &21.63&0.621&0.508
            &21.28&0.546&0.574\\

			\bottomrule
		\end{tabular}
		}
		\vspace{-15pt}
	\end{center}
\end{table*}

\section{Experiments}
\label{ExperimentalResults}
\subsection{Setting}

To verify the applicability and performance of the built dataset for evaluating NeRF methods, and meanwhile provide a baseline for NeRF-based tasks, we train and evaluate recent NeRF~\cite{mildenhall2020nerf}, NeRF++~\cite{zhang2020nerf++}, Mip-NeRF~\cite{barron2021mip}, Mip-NeRF 360~\cite{barron2022mip}, Mega-NeRF~\cite{turki2022mega} and Ref-NeRF~\cite{verbin2022ref} on our datasets.

\textbf{NeRF}~\cite{mildenhall2020nerf} presents the first continuous MLP-based neural network to represent the scene, that is able to synthesize semantic-consistent novel views by volume rendering.

\textbf{NeRF++}~\cite{zhang2020nerf++} separately models the foreground and background neural representations to address the challenge of modeling large-scale unbounded scenes.

\textbf{Mip-NeRF}~\cite{barron2021mip} reduces aliasing artifacts and improves NeRF’s~\cite{mildenhall2020nerf} ability to represent fine details, by rendering anti-aliased conical frustums instead of rays.

\textbf{Mip-NeRF 360}~\cite{barron2022mip} uses a non-linear scene parameterization, online distillation, and a distortion-based regularizer, to model and produce realistic synthesized views for unbounded real-world scenes.

\textbf{Mega-NeRF}~\cite{turki2022mega} proposes a framework for training large-scale 3D scenes by introducing a sparse structure and geometric clustering algorithm, to partition training pixels into different parallel NeRF submodules.


\textbf{Ref-NeRF}~\cite{verbin2022ref} improves the quality of appearance and normal in synthesized views of the scene, by a new parameterization and structuring of view-dependent outgoing radiance, as well as a regularizer on normal vectors.


\textbf{Implementations.} Since there is no official PyTorch implementation of NeRF~\cite{mildenhall2020nerf}, we use the widely recognized third-party implementation~\cite{pyNeRF}. But for other methods, we use the official implementation from GitHub.

In our dataset, reviewed posed images are numbered from 0 to $\#image-1$ under timing sequence, and test views are evenly sampled according to the view ID. That is we sample one from every eight for testing, and the rest are used as training views (i.e., for testing: 0, 8, 16, 24, ...). 
 
All training hyper-parameters follow the original paper's settings in our experiments. Each scene is trained on a single Nvidia V100 GPU device for around 6-33 hours, depending on the time complexity of each method, and about 32 V100 GPU devices are used in parallel.

\textbf{Evaluation Metrics.} To evaluate the performance of each method, we use three common metrics: Peak Signal-to-Noise Ratio (PSNR), Structural Similarity (SSIM)~\cite{wang2004ssim}, and LPIPS~\cite{zhang2018lpips}  on novel view synthesis. Higher PSNR and SSIM mean better performancence, while a lower LPIPS means better.

\subsection{Novel View Synthesis}

\begin{table*}[t]
	\begin{center}
		\centering
		\caption{More sub-benchmarks for novel view synthesis. We divide our dataset into subsets based on different scene types, camera trajectories, and lighting conditions, and provide sub-benchmarks under different settings. $\uparrow$ means the higher, the better.}
		\label{tab:subnovelviewsynthesis}
		\resizebox{1\textwidth}{!}{
		\begin{tabular}{cccccccccccccccccccc}
			\toprule
			\multirow{2}{*}{Scene ID}  & \multirow{2}{*}{Sub-benchmark} & \multicolumn{3}{c}{\textbf{NeRF}~\cite{mildenhall2020nerf}}& 
			\multicolumn{3}{c}{\textbf{NeRF++}~\cite{zhang2020nerf++}}& 
			\multicolumn{3}{c}{\textbf{Mip-NeRF}~\cite{barron2021mip}}&
			\multicolumn{3}{c}{\textbf{Mip-NeRF 360}~\cite{barron2022mip}}&
			\multicolumn{3}{c}{\textbf{Mega-NeRF}~\cite{turki2022mega}}& 
			\multicolumn{3}{c}{\textbf{Ref-NeRF}~\cite{xiangli2022bungeenerf}}\\
			 &  & PSNR$\uparrow$&SSIM$\uparrow$&LPIPS$\downarrow$& PSNR$\uparrow$&SSIM$\uparrow$&LPIPS$\downarrow$& PSNR$\uparrow$&SSIM$\uparrow$&LPIPS$\downarrow$&
			 PSNR$\uparrow$&SSIM$\uparrow$&LPIPS$\downarrow$&
			 PSNR$\uparrow$&SSIM$\uparrow$&LPIPS$\downarrow$& PSNR$\uparrow$&SSIM$\uparrow$&LPIPS$\downarrow$\\

		\cmidrule(l){1-1}  \cmidrule(l){2-2} \cmidrule(l){3-5} \cmidrule(l){6-8} \cmidrule(l){9-11} \cmidrule(l){12-14}\cmidrule(l){15-17} \cmidrule(l){18-20}

            1,4,7,8,11,13,22,26& \textbf{Buildings} 
            &17.32 & 0.501 & 0.605 		
            &22.24 & 0.644 & 0.528
            &17.82 & 0.546 & 0.636
            &22.85 & 0.704 & 0.444
            &21.05 & 0.650 & 0.511 		
            &22.30 & 0.633 & 0.541 \\
            
            2,5,12,14,15,18,19,28,29& \textbf{Small areas} 
            & 20.88 & 0.579 & 0.561
            & 21.66 & 0.621 & 0.514 		 
            & 19.44 & 0.545 & 0.588  
            & 24.89 & 0.742 & 0.358 
            & 22.90 & 0.668 & 0.486 
            & 20.56 & 0.553 & 0.582 \\
            
            3,8,9,10,20,27,31,32,33& \textbf{Cities} 
            &15.87 & 0.360 &0.637 		
            &19.92 & 0.496 &0.574 		
            &16.72 & 0.430 &0.650 	
            &20.44 & 0.548 &0.471 	
            &19.75 & 0.565 &0.523 	
            &20.33 & 0.492 &0.592 \\
            
            6,16,17,21,23,24,25,30& \textbf{Natural scenes} 
            &21.23 	&0.503 	&0.597 		
            &22.34 	&0.548 	&0.534 		
            &19.86 	&0.483 	&0.622 		
            &24.58 	&0.708 	&0.396 		
            &22.29 	&0.588 	&0.504 		
            &22.16 	&0.508 	&0.582 \\
            
            \midrule
            
            2,4,5,6,8,11,12,15,16,17,19,20,21,28,31& \textbf{Circles} 
            &21.08 	&0.573 	&0.559 		
            &22.00 	&0.609 	&0.508 		
            &19.80 	&0.545 	&0.581 		
            &23.81 	&0.713 	&0.386 		
            &21.74 	&0.647 	&0.483 		
            &21.53 	&0.564 	&0.556 \\
            
            3,7,9,13,14,18,22,23,24,25& \textbf{Lines} 
            &18.24 	&0.450 	&0.628 		
            &20.99 	&0.544 	&0.560 		
            &17.79 	&0.474 	&0.640 		
            &23.68 	&0.664 	&0.423 		
            &22.04 	&0.594 	&0.521 		
            &21.55 	&0.533 	&0.581 \\
            
            1,10,26,27,29,30,32,33& \textbf{Irregular} 
            &14.91 	&0.361 	&0.644 		
            &21.01 	&0.556 	&0.570 		
            &16.52 	&0.452 	&0.682 		
            &21.05 	&0.608 	&0.476 		
            &20.94 	&0.608 	&0.540 		
            &20.51 	&0.531 	&0.599 \\
            
            \midrule

            ALL-\{4, 11,31\} & \textbf{Day} &18.37 	&0.465 	&0.610 		
            &21.23 	&0.563 	&0.547 		
            &18.12 	&0.487 	&0.634 		
            &23.15 	&0.670 	&0.420 		
            &21.51 	&0.607 	&0.514 		
            &21.05 	&0.531 	&0.585 \\
            
            4, 11,31& \textbf{Night} 
            &22.24 	&0.680 	&0.499 		
            &23.71 	&0.707 	&0.455 		
            &21.17 	&0.646 	&0.517 		
            &22.61 	&0.703 	&0.406 		
            &22.89 	&0.765 	&0.452 		
            &23.62 	&0.707 	&0.468 \\

			\bottomrule
		\end{tabular}
		}
	    \vspace{-9pt}
	\end{center}
\end{table*}

\textbf{Benchmark.} To establish a benchmark for the large-scale outdoor novel view synthesis, we comprehensively evaluate and report quantitative performances of the above six state-of-the-art methods in our dataset, see Table ~\ref{tab:novelviewsynthesis}.

It can be seen that except for the failure of NeRF~\cite{mildenhall2020nerf} in 4 scenes (7, 26, 27, and 33), other results show that NeRF can synthesize reasonable novel views, which means that OMMO dataset can support various NeRF-based methods. NeRF++~\cite{zhang2020nerf++}, Mip-NeRF 360~\cite{barron2022mip}, Mega-NeRF~\cite{turki2022mega}, and Ref-NeRF~\cite{verbin2022ref} perform well on our dataset with an average PSNR of beyond 20, and can maintain the view consistency of each scene, see Figure~\ref{fig:exp}. Among them, Mip-NeRF 360~\cite{barron2022mip} can synthesize more realistic detailed texture features for large-scale scenes and its quantitative evaluation is more than 6 points higher than other methods on PSNR, SSIM, LPIPS. Our benchmarks are open to all NeRF-based methods, and we are also ready to evaluate newer large-scale scene NeRF methods once they are proposed.

In particular, we notice that most of the scenes where NeRF fails are based on irregular camera trajectories, which suggests that NeRF may be more suitable for scenes captured with stronger trajectory consistency constraints and more overlap (such as equidistant circular acquisitions). So we divide OMMO dataset into subsets according to the data types, and provide sub-benchmarks to study the most suitable setting for each method.

\begin{figure}[t]
	\centering
	\includegraphics[width=1\linewidth]{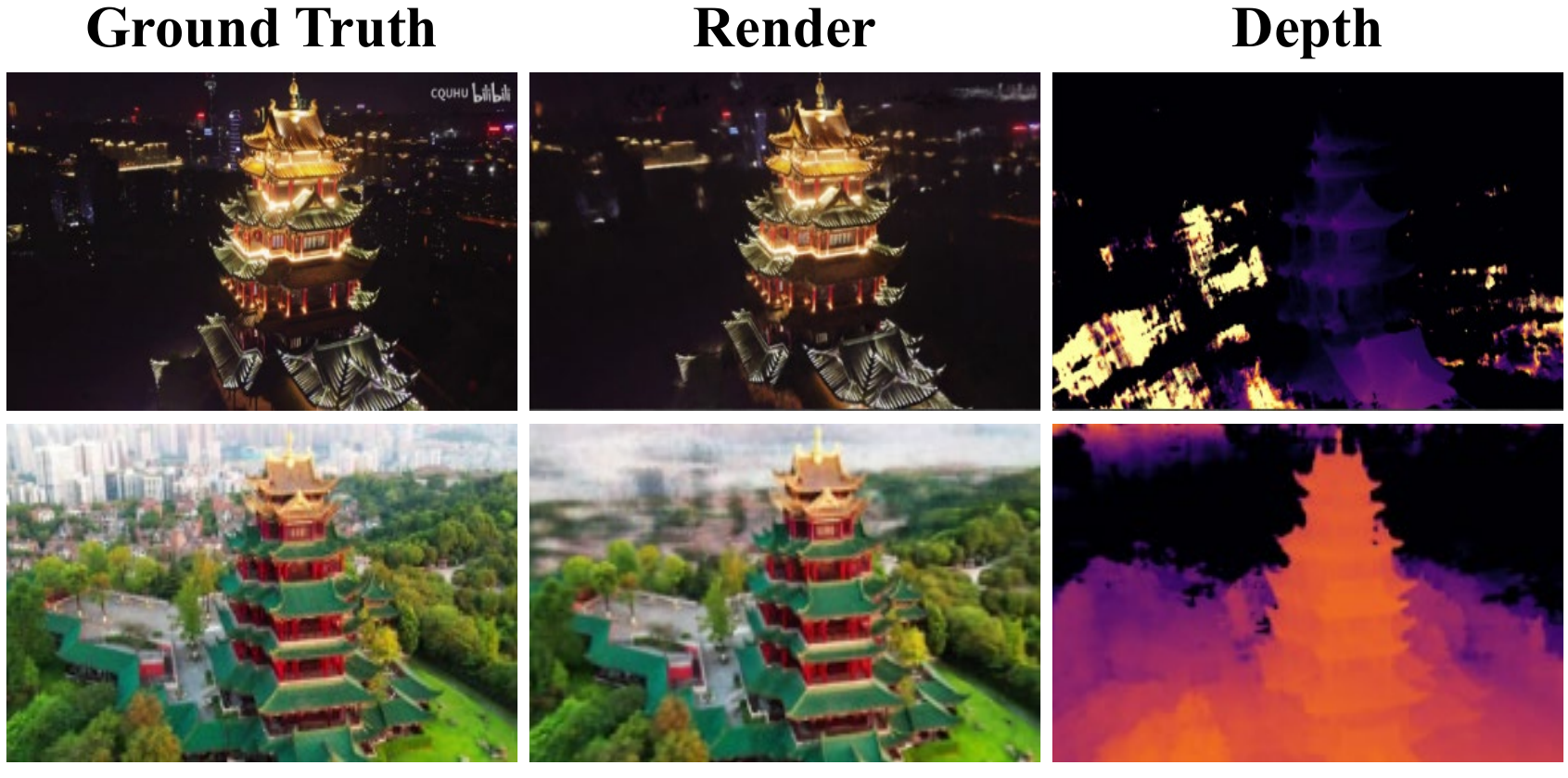}
	\vspace{-15pt}
	\caption{Different scans of the same scene during day and night. Both RGB and depth images are synthesized by Mega-NeRF~\cite{turki2022mega}.}
	\label{fig:day}
\end{figure}

\textbf{Sub-benchmark split by scene types.} According to different scales of urban and natural scenes, we propose 4 sub-benchmarks for buildings, small areas, cities and natural scenes. It can be seen from Table ~\ref{tab:subnovelviewsynthesis} that all methods perform worse in cities than in  smaller-scale subsets, i.e. buildings and small areas. These performance differences show that the large-scale scene implicit representation is still not as well resolved as for single objects or small scenes.

\textbf{Sub-benchmark split by camera tracks.} Circular camera trajectories tend to present better experiment performance than other types, especially irregular ones. This is because 360-degree views contain richer texture features from different angles, and the focus of views is overlapped to maintain the view consistency.

\begin{figure}[t]
	\centering
	\includegraphics[width=1\linewidth]{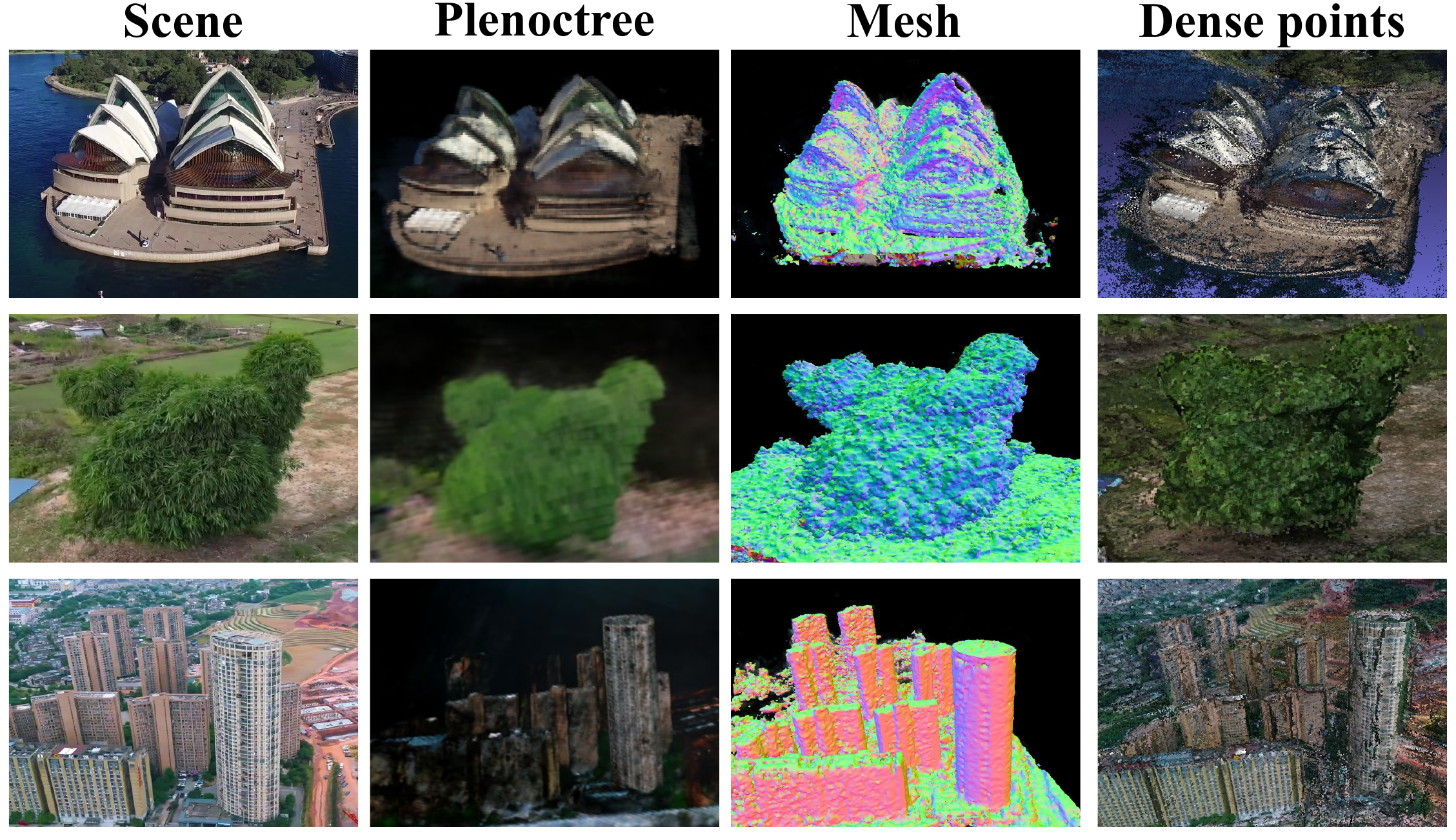}
	\vspace{-15pt}
	\caption{Examples of various scene representations from our dataset through different methods.}
	\label{fig:nor}
\end{figure}

\textbf{Sub-benchmark split by lighting conditions.} Intuitively, daytime scenes are richer in texture and easier to learn their representation than dark ones. However, we find that almost every method performs better on the low-light subset than on the normal-light subset. We visualize two different scans of the same scene generated by Mega-NeRF~\cite{turki2022mega} during day and night, as shown in Figure ~\ref{fig:day}. It is not difficult to see that in low light settings, the implicit network uses black areas to erase details when generating the RGB images, which reduces the synthesis difficulty and tricks the evaluation metrics, while synthesised poor depth map illustrates the network's incapacity to understand and represent the scene. So efficient low-light NeRF methods are urged to solve this problem. %


\subsection{Scene Representation}

To evaluate the performance of our dataset on surface or scene reconstruction tasks, we reconstruct scenes with different representations by using a variety of methods including implicit networks. Specifically, plenoctree, mesh, and dense points are provided by Mega-NeRF~\cite{turki2022mega}, Instant-NGP~\cite{mueller2022instant}, and Colmap~\cite{schonberger2016structure,schonberger2016pixelwise}, respectively. It can be seen from Figure~\ref{fig:nor} that neither the implicit network nor the feature matching reconstruction method can reconstruct the large scene finely. Theoretically, the advantage of implicit scene representation is that, the scene can be reconstructed with high resolution benefiting from the continuous representation. So the scene representation benchmarks of large-scale outdoor scenes based on NeRF are still to be build.
Please refer to the supplementary materials for more results.

\section{Discussion}

\textbf{Conclusion.}
We introduce a well-selected large-scale outdoor multi-modal fly-view dataset, OMMO, to address the problem of no widely-used benchmark for outdoor NeRF-based methods. The built OMMO surpasses the previous datasets in several key indicators such as quantity, quality and variety, by providing 33 real-world scenes with more than 14K posed images  and text description. 
With the help of our cost-effective data collection pipeline, it is easy to expand our dataset by continuously converting new internet videos into NeRF-purpose training data.
We provide benchmarks on multiple tasks such as novel view synthesis, implicit scene representations, and multi-modal synthesis by evaluating existing methods. Experiments show that our dataset can well support mainstream NeRF-based tasks.

\textbf{Limitation.} 
Scenes with low-illumination, rain and fog are still few in the current dataset due to the limited calibration and reconstruction ability of COLMAP~\cite{schonberger2016structure,schonberger2016pixelwise}. We will try more reconstruction and calibration methods to solve this problem. Meanwhile, we are continuing to expand our dataset thanks to our cost-effective pipeline.

{\small
\bibliographystyle{ieee_fullname}
\bibliography{egbib}

\begin{thebibliography}{10}\itemsep=-1pt

\bibitem{agarwal2011building}
Sameer Agarwal, Yasutaka Furukawa, Noah Snavely, Ian Simon, Brian Curless,
  Steven~M Seitz, and Richard Szeliski.
\newblock Building rome in a day.
\newblock {\em Communications of the ACM}, 54(10):105--112, 2011.

\bibitem{barron2021mip}
Jonathan~T Barron, Ben Mildenhall, Matthew Tancik, Peter Hedman, Ricardo
  Martin-Brualla, and Pratul~P Srinivasan.
\newblock Mip-nerf: A multiscale representation for anti-aliasing neural
  radiance fields.
\newblock In {\em Proceedings of the IEEE/CVF International Conference on
  Computer Vision}, pages 5855--5864, 2021.

\bibitem{barron2022mip}
Jonathan~T Barron, Ben Mildenhall, Dor Verbin, Pratul~P Srinivasan, and Peter
  Hedman.
\newblock Mip-nerf 360: Unbounded anti-aliased neural radiance fields.
\newblock In {\em Proceedings of the IEEE/CVF Conference on Computer Vision and
  Pattern Recognition}, pages 5470--5479, 2022.

\bibitem{bozic2021transformerfusion}
Aljaz Bozic, Pablo Palafox, Justus Thies, Angela Dai, and Matthias Nie{\ss}ner.
\newblock Transformerfusion: Monocular rgb scene reconstruction using
  transformers.
\newblock {\em Advances in Neural Information Processing Systems},
  34:1403--1414, 2021.

\bibitem{chen2022geoaug}
Di Chen, Yu Liu, Lianghua Huang, Bin Wang, and Pan Pan.
\newblock Geoaug: Data augmentation for few-shot nerf with geometry
  constraints.
\newblock In {\em European Conference on Computer Vision}, pages 322--337.
  Springer, 2022.

\bibitem{crandall2011discrete}
David Crandall, Andrew Owens, Noah Snavely, and Dan Huttenlocher.
\newblock Discrete-continuous optimization for large-scale structure from
  motion.
\newblock In {\em CVPR 2011}, pages 3001--3008. IEEE, 2011.

\bibitem{dai2017scannet}
Angela Dai, Angel~X Chang, Manolis Savva, Maciej Halber, Thomas Funkhouser, and
  Matthias Nie{\ss}ner.
\newblock Scannet: Richly-annotated 3d reconstructions of indoor scenes.
\newblock In {\em Proceedings of the IEEE conference on computer vision and
  pattern recognition}, pages 5828--5839, 2017.

\bibitem{deng2009imagenet}
Jia Deng, Wei Dong, Richard Socher, Li-Jia Li, Kai Li, and Li Fei-Fei.
\newblock Imagenet: A large-scale hierarchical image database.
\newblock In {\em 2009 IEEE conference on computer vision and pattern
  recognition}, pages 248--255. Ieee, 2009.

\bibitem{deng2022depth}
Kangle Deng, Andrew Liu, Jun-Yan Zhu, and Deva Ramanan.
\newblock Depth-supervised nerf: Fewer views and faster training for free.
\newblock In {\em Proceedings of the IEEE/CVF Conference on Computer Vision and
  Pattern Recognition}, pages 12882--12891, 2022.

\bibitem{pyNeRF}
Github.
\newblock nerf-pytorch.
\newblock \url{https://github.com/yenchenlin/nerf-pytorch}.

\bibitem{grubinger2006iapr}
Michael Grubinger, Paul Clough, Henning M{\"u}ller, and Thomas Deselaers.
\newblock The iapr tc-12 benchmark: A new evaluation resource for visual
  information systems.
\newblock In {\em International workshop ontoImage}, volume~2, 2006.

\bibitem{guo2022neural}
Haoyu Guo, Sida Peng, Haotong Lin, Qianqian Wang, Guofeng Zhang, Hujun Bao, and
  Xiaowei Zhou.
\newblock Neural 3d scene reconstruction with the manhattan-world assumption.
\newblock In {\em Proceedings of the IEEE/CVF Conference on Computer Vision and
  Pattern Recognition}, pages 5511--5520, 2022.

\bibitem{heinly2015reconstructing}
Jared Heinly, Johannes~L Schonberger, Enrique Dunn, and Jan-Michael Frahm.
\newblock Reconstructing the world* in six days*(as captured by the yahoo 100
  million image dataset).
\newblock In {\em Proceedings of the IEEE conference on computer vision and
  pattern recognition}, pages 3287--3295, 2015.

\bibitem{jensen2014dtu}
Rasmus Jensen, Anders Dahl, George Vogiatzis, Engin Tola, and Henrik Aan{\ae}s.
\newblock Large scale multi-view stereopsis evaluation.
\newblock In {\em Proceedings of the IEEE conference on computer vision and
  pattern recognition}, pages 406--413, 2014.

\bibitem{johari2022geonerf}
Mohammad~Mahdi Johari, Yann Lepoittevin, and Fran{\c{c}}ois Fleuret.
\newblock Geonerf: Generalizing nerf with geometry priors.
\newblock In {\em Proceedings of the IEEE/CVF Conference on Computer Vision and
  Pattern Recognition}, pages 18365--18375, 2022.

\bibitem{kania2022conerf}
Kacper Kania, Kwang~Moo Yi, Marek Kowalski, Tomasz Trzci{\'n}ski, and Andrea
  Tagliasacchi.
\newblock Conerf: Controllable neural radiance fields.
\newblock In {\em Proceedings of the IEEE/CVF Conference on Computer Vision and
  Pattern Recognition}, pages 18623--18632, 2022.

\bibitem{kim2021representative}
Hanul Kim, Su-Min Choi, Chang-Su Kim, and Yeong~Jun Koh.
\newblock Representative color transform for image enhancement.
\newblock In {\em Proceedings of the IEEE/CVF International Conference on
  Computer Vision}, pages 4459--4468, 2021.

\bibitem{knapitsch2017tanks}
Arno Knapitsch, Jaesik Park, Qian-Yi Zhou, and Vladlen Koltun.
\newblock Tanks and temples: Benchmarking large-scale scene reconstruction.
\newblock {\em ACM Transactions on Graphics (ToG)}, 36(4):1--13, 2017.

\bibitem{kulhanek2022viewformer}
Jon{\'a}{\v{s}} Kulh{\'a}nek, Erik Derner, Torsten Sattler, and Robert
  Babu{\v{s}}ka.
\newblock Viewformer: Nerf-free neural rendering from few images using
  transformers.
\newblock {\em arXiv preprint arXiv:2203.10157}, 2022.

\bibitem{legros2020robust}
Quentin Legros, Juli{\'a}n Tachella, Rachael Tobin, Aongus McCarthy, Sylvain
  Meignen, Gerald~S Buller, Yoann Altmann, Stephen McLaughlin, and Michael~E
  Davies.
\newblock Robust 3d reconstruction of dynamic scenes from single-photon lidar
  using beta-divergences.
\newblock {\em IEEE Transactions on Image Processing}, 30:1716--1727, 2020.

\bibitem{lin2014microsoft}
Tsung-Yi Lin, Michael Maire, Serge Belongie, James Hays, Pietro Perona, Deva
  Ramanan, Piotr Doll{\'a}r, and C~Lawrence Zitnick.
\newblock Microsoft coco: Common objects in context.
\newblock In {\em European conference on computer vision}, pages 740--755.
  Springer, 2014.

\bibitem{liu2021urbanscene3d}
Yilin Liu, Fuyou Xue, and Hui Huang.
\newblock Urbanscene3d: A large scale urban scene dataset and simulator.
\newblock {\em arXiv preprint arXiv:2107.04286}, 2021.

\bibitem{martin2021wild}
Ricardo Martin-Brualla, Noha Radwan, Mehdi~SM Sajjadi, Jonathan~T Barron,
  Alexey Dosovitskiy, and Daniel Duckworth.
\newblock Nerf in the wild: Neural radiance fields for unconstrained photo
  collections.
\newblock In {\em Proceedings of the IEEE/CVF Conference on Computer Vision and
  Pattern Recognition}, pages 7210--7219, 2021.

\bibitem{mildenhall2020nerf}
Ben Mildenhall, Pratul~P. Srinivasan, Matthew Tancik, Jonathan~T. Barron, Ravi
  Ramamoorthi, and Ren Ng.
\newblock Nerf: Representing scenes as neural radiance fields for view
  synthesis.
\newblock In {\em ECCV}, 2020.

\bibitem{mueller2022instant}
Thomas M\"uller, Alex Evans, Christoph Schied, and Alexander Keller.
\newblock Instant neural graphics primitives with a multiresolution hash
  encoding.
\newblock {\em ACM Trans. Graph.}, 41(4):102:1--102:15, July 2022.

\bibitem{niemeyer2022regnerf}
Michael Niemeyer, Jonathan~T Barron, Ben Mildenhall, Mehdi~SM Sajjadi, Andreas
  Geiger, and Noha Radwan.
\newblock Regnerf: Regularizing neural radiance fields for view synthesis from
  sparse inputs.
\newblock In {\em Proceedings of the IEEE/CVF Conference on Computer Vision and
  Pattern Recognition}, pages 5480--5490, 2022.

\bibitem{oechsle2021unisurf}
Michael Oechsle, Songyou Peng, and Andreas Geiger.
\newblock Unisurf: Unifying neural implicit surfaces and radiance fields for
  multi-view reconstruction.
\newblock In {\em Proceedings of the IEEE/CVF International Conference on
  Computer Vision}, pages 5589--5599, 2021.

\bibitem{radford2021learning}
Alec Radford, Jong~Wook Kim, Chris Hallacy, Aditya Ramesh, Gabriel Goh,
  Sandhini Agarwal, Girish Sastry, Amanda Askell, Pamela Mishkin, Jack Clark,
  et~al.
\newblock Learning transferable visual models from natural language
  supervision.
\newblock In {\em International Conference on Machine Learning}, pages
  8748--8763. PMLR, 2021.

\bibitem{ramon2021h3d}
Eduard Ramon, Gil Triginer, Janna Escur, Albert Pumarola, Jaime Garcia, Xavier
  Giro-i Nieto, and Francesc Moreno-Noguer.
\newblock H3d-net: Few-shot high-fidelity 3d head reconstruction.
\newblock In {\em Proceedings of the IEEE/CVF International Conference on
  Computer Vision}, pages 5620--5629, 2021.

\bibitem{sanabria2018how2}
Ramon Sanabria, Ozan Caglayan, Shruti Palaskar, Desmond Elliott, Lo\"ic
  Barrault, Lucia Specia, and Florian Metze.
\newblock {How2:} a large-scale dataset for multimodal language understanding.
\newblock In {\em Proceedings of the Workshop on Visually Grounded Interaction
  and Language (ViGIL)}. NeurIPS, 2018.

\bibitem{schonberger2016structure}
Johannes~L Schonberger and Jan-Michael Frahm.
\newblock Structure-from-motion revisited.
\newblock In {\em Proceedings of the IEEE conference on computer vision and
  pattern recognition}, pages 4104--4113, 2016.

\bibitem{schonberger2016pixelwise}
Johannes~L Sch{\"o}nberger, Enliang Zheng, Jan-Michael Frahm, and Marc
  Pollefeys.
\newblock Pixelwise view selection for unstructured multi-view stereo.
\newblock In {\em European conference on computer vision}, pages 501--518.
  Springer, 2016.

\bibitem{talebi2018nima}
Hossein Talebi and Peyman Milanfar.
\newblock Nima: Neural image assessment.
\newblock {\em IEEE transactions on image processing}, 27(8):3998--4011, 2018.

\bibitem{tancik2022block}
Matthew Tancik, Vincent Casser, Xinchen Yan, Sabeek Pradhan, Ben Mildenhall,
  Pratul~P Srinivasan, Jonathan~T Barron, and Henrik Kretzschmar.
\newblock Block-nerf: Scalable large scene neural view synthesis.
\newblock In {\em Proceedings of the IEEE/CVF Conference on Computer Vision and
  Pattern Recognition}, pages 8248--8258, 2022.

\bibitem{turki2022mega}
Haithem Turki, Deva Ramanan, and Mahadev Satyanarayanan.
\newblock Mega-nerf: Scalable construction of large-scale nerfs for virtual
  fly-throughs.
\newblock In {\em Proceedings of the IEEE/CVF Conference on Computer Vision and
  Pattern Recognition}, pages 12922--12931, 2022.

\bibitem{verbin2022ref}
Dor Verbin, Peter Hedman, Ben Mildenhall, Todd Zickler, Jonathan~T Barron, and
  Pratul~P Srinivasan.
\newblock Ref-nerf: Structured view-dependent appearance for neural radiance
  fields.
\newblock In {\em 2022 IEEE/CVF Conference on Computer Vision and Pattern
  Recognition (CVPR)}, pages 5481--5490. IEEE, 2022.

\bibitem{wang2021neus}
Peng Wang, Lingjie Liu, Yuan Liu, Christian Theobalt, Taku Komura, and Wenping
  Wang.
\newblock Neus: Learning neural implicit surfaces by volume rendering for
  multi-view reconstruction.
\newblock {\em Advances in Neural Information Processing Systems},
  34:27171--27183, 2021.

\bibitem{wang2004ssim}
Zhou Wang, Alan~C Bovik, Hamid~R Sheikh, and Eero~P Simoncelli.
\newblock Image quality assessment: from error visibility to structural
  similarity.
\newblock {\em IEEE transactions on image processing}, 13(4):600--612, 2004.

\bibitem{xiangli2022bungeenerf}
Yuanbo Xiangli, Linning Xu, Xingang Pan, Nanxuan Zhao, Anyi Rao, Christian
  Theobalt, Bo Dai, and Dahua Lin.
\newblock Bungeenerf: Progressive neural radiance field for extreme multi-scale
  scene rendering.
\newblock In {\em The European Conference on Computer Vision (ECCV)}, volume~2,
  2022.

\bibitem{yang2022recursive}
Guo-Wei Yang, Wen-Yang Zhou, Hao-Yang Peng, Dun Liang, Tai-Jiang Mu, and
  Shi-Min Hu.
\newblock Recursive-nerf: An efficient and dynamically growing nerf.
\newblock {\em IEEE Transactions on Visualization and Computer Graphics}, 2022.

\bibitem{yao2020blendedmvs}
Yao Yao, Zixin Luo, Shiwei Li, Jingyang Zhang, Yufan Ren, Lei Zhou, Tian Fang,
  and Long Quan.
\newblock Blendedmvs: A large-scale dataset for generalized multi-view stereo
  networks.
\newblock In {\em Proceedings of the IEEE/CVF Conference on Computer Vision and
  Pattern Recognition}, pages 1790--1799, 2020.

\bibitem{yariv2021volume}
Lior Yariv, Jiatao Gu, Yoni Kasten, and Yaron Lipman.
\newblock Volume rendering of neural implicit surfaces.
\newblock {\em Advances in Neural Information Processing Systems},
  34:4805--4815, 2021.

\bibitem{yin2022coordinates}
Fukun Yin, Wen Liu, Zilong Huang, Pei Cheng, Tao Chen, and Gang YU.
\newblock Coordinates are not lonely--codebook prior helps implicit neural 3d
  representations.
\newblock {\em arXiv preprint arXiv:2210.11170}, 2022.

\bibitem{zhang2020nerf++}
Kai Zhang, Gernot Riegler, Noah Snavely, and Vladlen Koltun.
\newblock Nerf++: Analyzing and improving neural radiance fields.
\newblock {\em arXiv preprint arXiv:2010.07492}, 2020.

\bibitem{zhang2018lpips}
Richard Zhang, Phillip Isola, Alexei~A Efros, Eli Shechtman, and Oliver Wang.
\newblock The unreasonable effectiveness of deep features as a perceptual
  metric.
\newblock In {\em Proceedings of the IEEE conference on computer vision and
  pattern recognition}, pages 586--595, 2018.

\end{thebibliography}
}

\clearpage
\onecolumn
\section*{\hfil {\LARGE Appendix}\hfil}
\vspace{50pt}
\renewcommand\thesection{\Alph{section}}
\setcounter{section}{0}
\setcounter{section}{0}
\setcounter{figure}{0}
\setcounter{table}{0}

In this supplementary material, we provide the appendix section and a supplemental video to better understand our database and benchmarks. This appendix involves more qualitative or experimental results (\cref{supp:addE}), details of our dataset generation method (\cref{supp:method}), and dataset analysis (\cref{supp:datasetAnaysis}). The supplemental video contains a brief introduction to our dataset, some examples in detail, and more comprehensive synthesis results in surrounding views or progressive views. 


\section{More Qualitative Results}
\label{supp:moreQR}

\begin{figure}[b]
	\centering
	\includegraphics[width=1\linewidth]{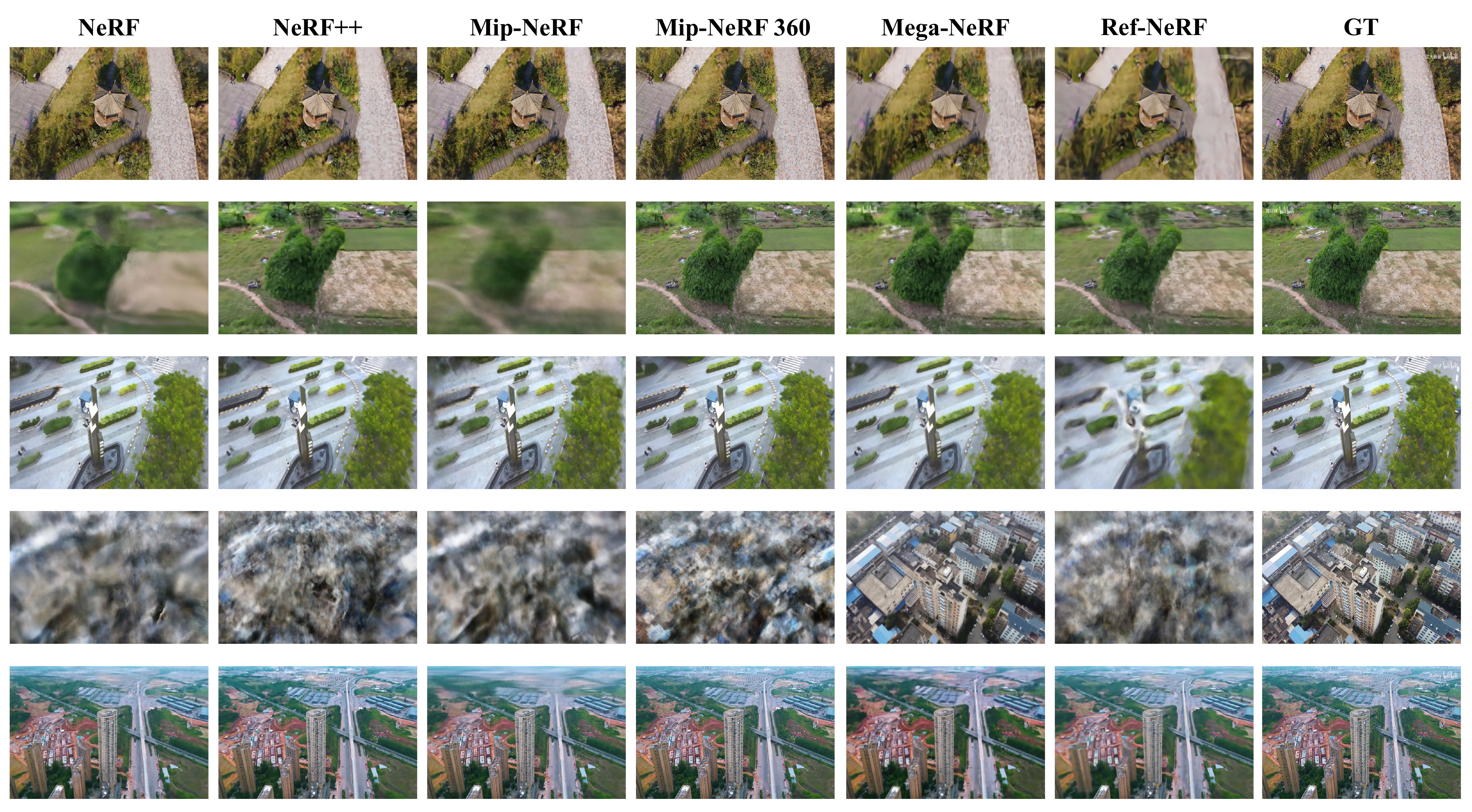}
	Part 1\,/\,2
	\caption{More qualitative visualization results for novel view synthesis (zoom-in for the best of views) on our OMMO dataset.}
	\label{fig:supp_nvs}
\end{figure}

\begin{figure}[t]
	\centering
    \includegraphics[width=1\linewidth]{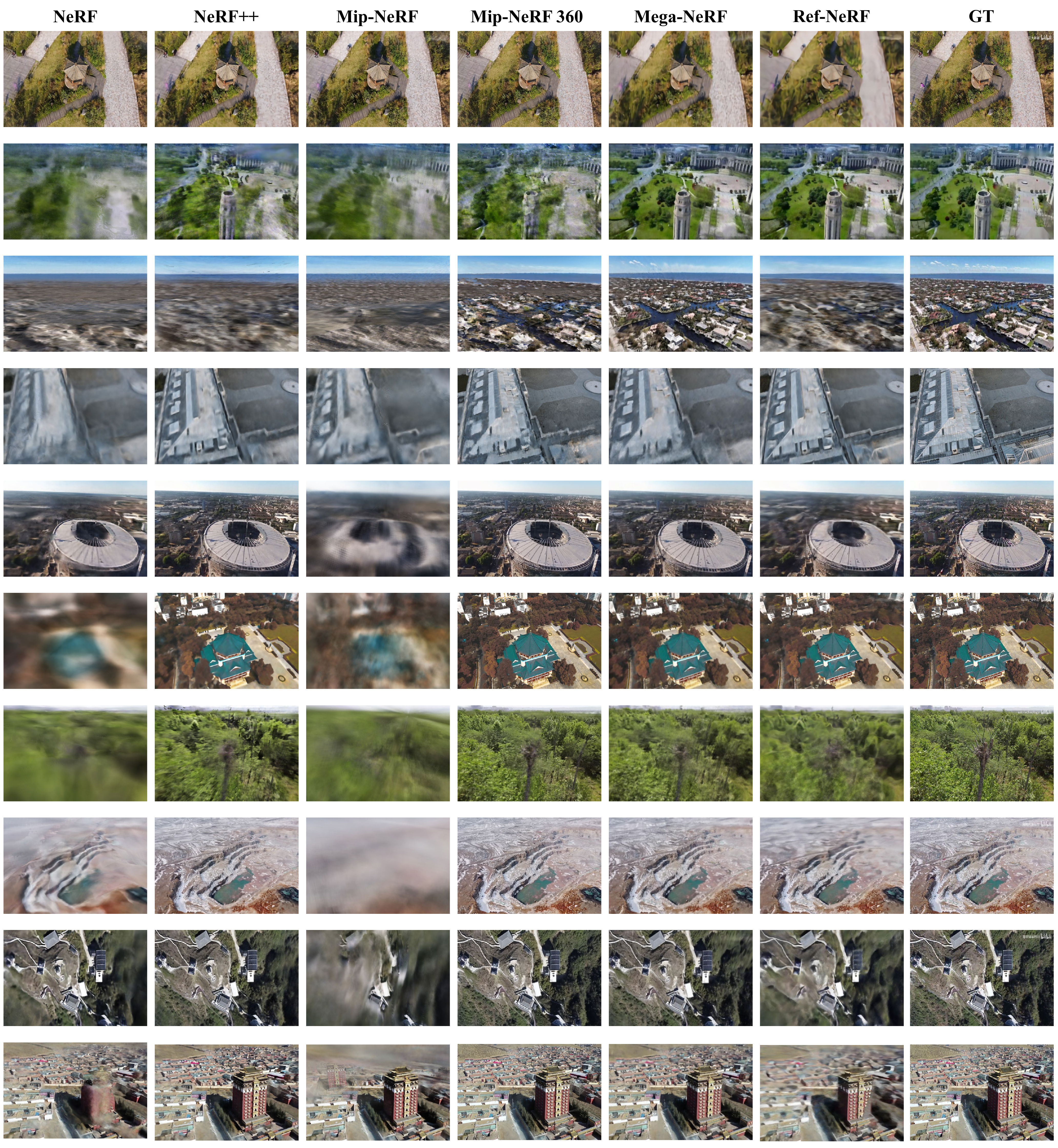}
    Part 2\,/\,2
	\caption*{Figure 1. More qualitative visualization results for novel view synthesis (zoom-in for the best of views) on our OMMO dataset.}
\end{figure}

\subsection{Novel View Synthesis}
\label{supp:moreQR_nvs}

In our main manuscript, we can only provide the visualization results of five scenes due to the length limitation. In this section, more qualitative results are presented to demonstrate the novel view synthesis ability of each method, see Fig.\ref{fig:supp_nvs}.

\clearpage

\subsection{Scene Representation}
\label{supp:moreQR_sr}

\begin{figure}[h]
	\centering
    \includegraphics[width=1\linewidth]{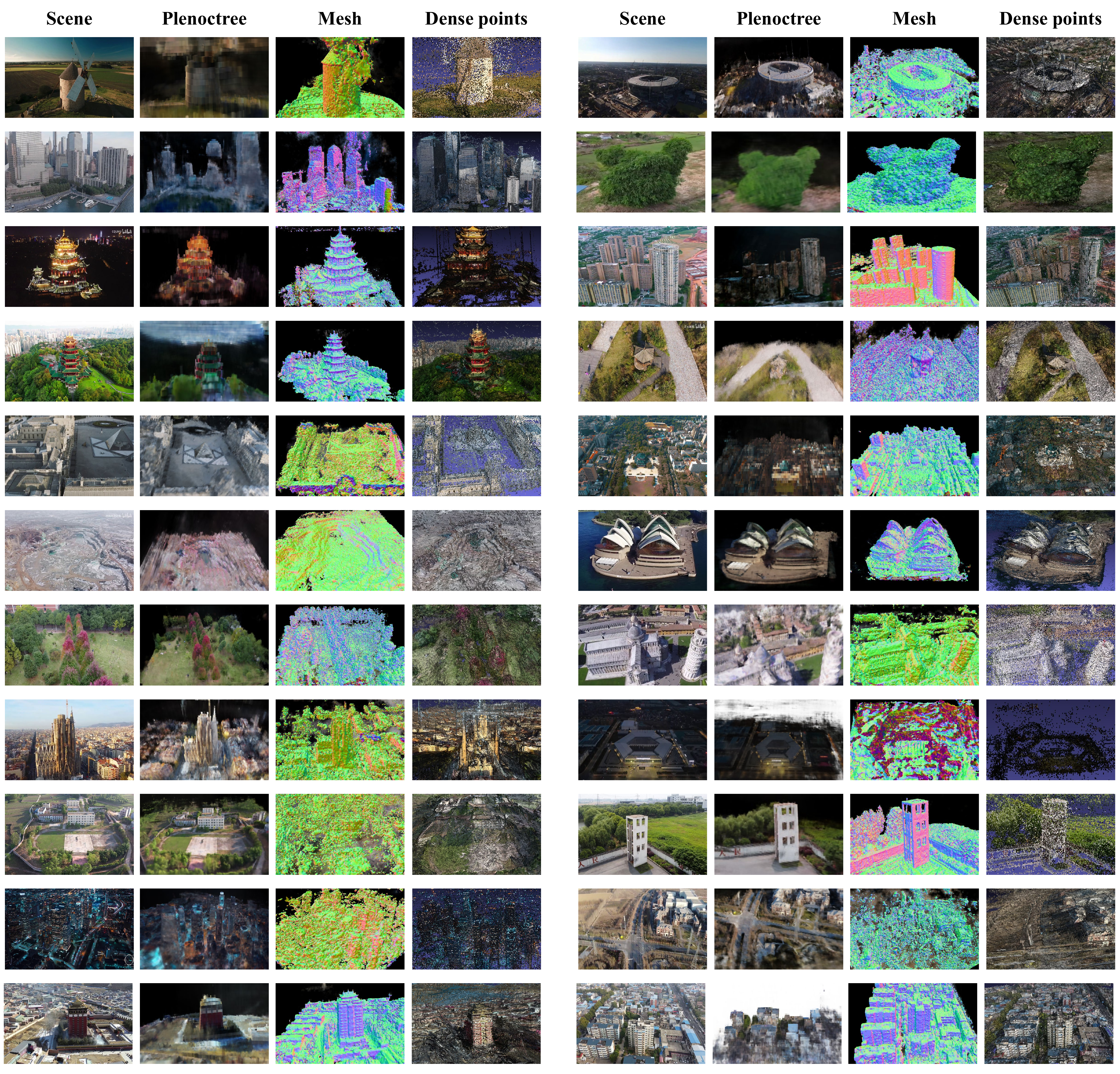}
	\caption{More qualitative visualization results for various scene representations (zoom-in for the best of views) through the state-of-the-art methods on the OMMO dataset.}
	\label{fig:suppnor}
\end{figure}

To further demonstrate that our OMMO dataset can well support surface or scene reconstruction tasks including NeRF-based methods, we visualize more shape results by various representations in Fig.~\ref{fig:suppnor}. Among them, plenoctree, mesh, and dense points are provided by Mega-NeRF~\cite{turki2022mega}, InstantNGP~\cite{mueller2022instant}, and Colmap~\cite{schonberger2016structure,schonberger2016pixelwise}, respectively.

\clearpage

\section{More Experiments}
\label{supp:addE}

\begin{figure}[h]
	\centering
	\includegraphics[width=1\linewidth]{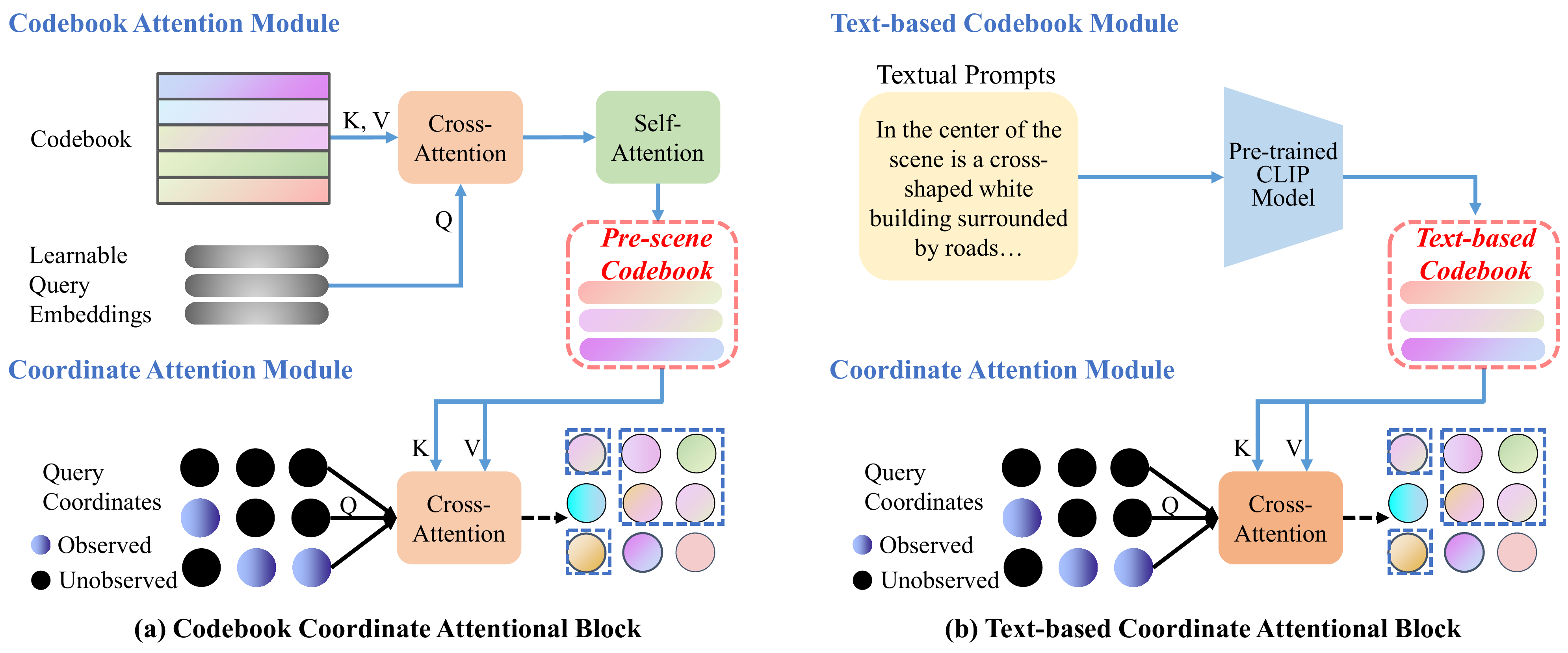}
	\caption{Structure of coordinate attentional blocks. The left sub-figure (a) is the CoCo-INR's~\cite{yin2022coordinates} codebook coordinate attentional block, which extracts image features related to the current scene from the prior by codebook attention module to form a pre-scene codebook. The right sub-figure (b) is our text-based coordinate attentional block, which obtains the scene-related text-based codebook by encoding textual prompts of each scene. Both inject scene-related features into each coordinate through the coordinate attention module.}
	\label{fig:suppmethod}
\end{figure}

\textbf{Multi-modal NeRF Synthesis.} Since there is no available NeRF-based method for text-assisted fidelity novel view synthesis, inspired by CoCo-INR~\cite{yin2022coordinates}, we replace its image-based pre-scene codebook with text-based codebook and apply it in NeRF~\cite{mildenhall2020nerf} and CoCo-INR~\cite{yin2022coordinates} as our benchmark. Specifically, we apply a text-based attentional coordinate module in front of the last MLP layer of volume rendering network, where the text-based codebook is encoded by our textual prompts through the pre-trained CLIP~\cite{radford2021learning} model, see Fig.~\ref{fig:suppmethod}. 

\begin{table*}[h]
	\begin{center}
		\centering
		\caption{Performance comparison of with or without textual prompts for novel view synthesis on our OMMO dataset. We report the results of each method and the performance improvement after injecting textual prompts. $\uparrow$ means the higher, the better.}
		\label{tab:suppmultim}
		\resizebox{0.6\textwidth}{!}{
		\begin{tabular}{c|ccc|ccc|ccc}
			\toprule
			\multirow{2}{*}{Mtehod} & 
			\multicolumn{3}{c|}{Without Textual Prompts}&
			\multicolumn{3}{c|}{With Textual Prompts}&
			\multicolumn{3}{c}{\textbf{Improvement (\%) $\uparrow$}}\\
			 &PSNR$\uparrow$&SSIM$\uparrow$&LPIPS$\downarrow$& PSNR$\uparrow$&SSIM$\uparrow$&LPIPS$\downarrow$&PSNR&SSIM&LPIPS\\

            \midrule
            \textbf{NeRF~\cite{mildenhall2020nerf}}
            &18.72 &0.484 &0.600
            &19.01&0.500&0.591
            &\textbf{1.5}&\textbf{3.2}&\textbf{1.5}\\
            \textbf{CoCo-INR~\cite{yin2022coordinates}}
            &16.80&0.489&0.681
            &16.97&0.490&0.678
            &\textbf{1.0}&\textbf{0.2}&\textbf{0.4}\\
			\bottomrule
		\end{tabular}
		}
	\end{center}
\end{table*}

It can be seen from Table~\ref{tab:suppmultim}, even without a well-designed module for injecting text information, the performance of both NeRF~\cite{mildenhall2020nerf} and CoCo-INR~\cite{yin2022coordinates} methods have improved. Since the textual prompts contain more global features about rich geometry or appearance information, which are shared by different views in the scene to guarantee the network to generate view-consistency results. We hope to inspire more image-text multi-modal NeRF methods to synthesize photo-realistic rendering results and decent geometry by exploring effective ways to make full use of textual prompts. The benchmark on each scene and the sub-benchmarks on different scene types are shown in Table~\ref{tab:suppmultim1} and Table~\ref{tab:suppmultim2}.

\begin{table*}[t]
	\begin{center}
		\centering
		\caption{Benchmark for multi-modal NeRF synthesis. We present the performance of text-assisted novel view synthesis based on existing methods on our OMMO dataset. $\uparrow$ means the higher, the better.}
		\label{tab:suppmultim1}
		\resizebox{1\textwidth}{!}{
		\begin{tabular}{c|ccc|ccc|ccc|ccc|ccc}
			\toprule
			\multirow{2}{*}{Scene ID}  & Scene & Camera  & Lighting & 
			\multicolumn{3}{c|}{\textbf{NeRF}~\cite{mildenhall2020nerf} w/o Prompts}& 
			\multicolumn{3}{c|}{\textbf{NeRF}~\cite{mildenhall2020nerf} w/ Prompts}& 
			\multicolumn{3}{c|}{\textbf{CoCo-INR}~\cite{yin2022coordinates} w/o Prompts}&
			\multicolumn{3}{c}{\textbf{CoCo-INR}~\cite{yin2022coordinates} w/ Prompts}\\
			
			 & Types & Tracks &Conditions & PSNR$\uparrow$&SSIM$\uparrow$&LPIPS$\downarrow$& PSNR$\uparrow$&SSIM$\uparrow$&LPIPS$\downarrow$& 
			 PSNR$\uparrow$&SSIM$\uparrow$&LPIPS$\downarrow$& PSNR$\uparrow$&SSIM$\uparrow$&LPIPS$\downarrow$\\

            \midrule

			1 & Buildings & Irregular & Day 
			&16.93&0.369&0.744
			& 16.89 & 0.366 & 0.729
			& 14.31  & 0.432  &0.788 
			& 14.81  & 0.431  &  0.785 \\
            2 & Small area & Circles & Day 
            &15.31&0.442&0.694 
            & 15.61 & 0.465 & 0.711
			& 16.04  & 0.597  &0.632 
			&16.25   & 0.597  & 0.626  \\
            3 & Citys & Lines & Day 
            &14.38 &0.278&0.556 
            & 14.42 & 0.277 &0.573
			& 15.59 &0.485   &0.616 
			& 16.62  &  0.509 & 0.585  \\
            4 & Buildings & Circle & Night 
            &25.39 &0.859 &0.431 
            & 24.94 &0.851  &0.425
			& 21.61  & 0.876  &0.480 
			& 21.87  & 0.879  & 0.481  \\
            5 & Small area & Circles & Day 
            &22.26 &0.670 &0.531 
            & 21.31 & 0.652 &0.564
			& 18.16  & 0.657  &0.597 
			& 20.08  & 0.675  & 0.573  \\
            6 & Natural scenes & Circles & Day 
            &24.09 &0.679 &0.504 
            & 23.78 & 0.655 &0.535
			& 19.65   & 0.630   &0.576 
			&  19.39 & 0.627  & 0.578  \\
            7 & Buildings & Lines & Day 
            &5.36 &0.166 &0.747 
            & 6.25 & 0.183 &0.697
			& 16.53  & 0.628  &0.679 
			& 15.38  &0.567   & 0.654  \\
            8 & Citys & Circle & Day 
            &21.14 &0.496 &0.594 
            & 21.55 & 0.510 &0.571
			& 16.94  & 0.413  &0.687 
			& 16.57  & 0.407  & 0.704  \\
            9 & Citys & Lines & Day 
            &14.92&0.344&0.744
            & 15.02 &  0.345&0.749
			& 13.70  & 0.340  &0.773 
			& 13.68  &  0.340 & 0.765  \\
            10 & Citys & Irregular & Day 
            &22.26 &0.550 &0.626 
            & 22.44 & 0.551 &0.624
			& 18.81  & 0.536  &0.694 
			&  18.62 &  0.535 &  0.693 \\
            11 & Buildings & Circles & Night 
            &22.36 &0.816 &0.420 
            & 22.58 & 0.820 &0.412
			& 17.08  & 0.746  &0.494 
			& 17.35  & 0.747  &  0.491 \\
            12 & Small area & Circles & Day 
            &22.41 &0.594 &0.533
            & 22.80 & 0.608 & 0.512
			& 17.87  & 0.475  &0.658 
			& 17.81  & 0.473  & 0.659  \\
            13 & Buildings & Lines & Day 
            &22.27 &0.592 &0.608 
            & 23.12 & 0.619 &0.576
			& 16.55  & 0.532  &0.698 
			& 17.02  & 0.542  & 0.671  \\
            14 & Small area & Lines & Day 
            &19.85 &0.554 &0.569 
            & 20.73 & 0.591 &0.534
			& 15.44  & 0.485  &0.663 
			& 15.19  & 0.482  & 0.665  \\
            15 & Small area & Circles & Day 
            &20.35&0.527&0.552 
            & 20.70 & 0.549 &0.533
			& 16.37  &0.407   &0.702 
			&16.45   & 0.411  & 0.689  \\
            16 & Natural scenes & Circles & Day 
            &17.86&0.397&0.631
            & 17.53 &0.362  &0.647
			& 15.37  &  0.384 &0.633 
			&15.24   &  0.376 & 0.640  \\
            17 & Natural scenes & Circles & Day 
            &22.02&0.571&0.610
            & 22.23 & 0.575 &0.596
			& 20.52  & 0.575  &0.619 
			& 19.38  & 0.527  & 0.648  \\
            18 & Small area & Lines & Day 
            &26.06 &0.754 &0.428 
            & 26.48 & 0.770 &0.402
			& 17.31  & 0.527  &0.658 
			& 17.35  &  0.532 & 0.664  \\
            19 & Small area & Circles & Day 
            &14.20&0.399&0.726
            & 14.19 & 0.397 &0.720
			& 15.41  &  0.388 &0.701 
			& 15.82  & 0.413  & 0.694  \\
            20 & Citys & Circles & Day 
            &22.84 &0.613 &0.499 
            & 23.30 & 0.636 &0.465
			& 18.28  & 0.434  &0.676 
			& 18.09  & 0.431  & 0.685  \\
            21 & Natural scenes & Circles & Day 
            &22.59&0.514&0.532
            & 22.99 & 0.541 &0.508
			& 17.08  &0.358   &0.744 
			& 17.28  & 0.359  & 0.720  \\
            22 & Buildings & Lines & Day 
            &16.53 &0.466 &0.733
            & 20.404 &  0.539&0.598
			&14.86   & 0.408  &0.759 
			& 14.73  &  0.406 & 0.772  \\
            23 & Natural scenes & Lines & Day 
            &18.99&0.405&0.669
            & 19.09 & 0.405 &0.671
			& 17.57  & 0.335  &0.673 
			&  17.43 & 0.332  & 0.701  \\
            24 & Natural scenes & Lines & Day 
            &19.32&0.386&0.696
            & 18.52 & 0.379 &0.708
			& 18.63  & 0.347  &0.765 
			&  18.27 & 0.341  & 0.814  \\
            25 & Natural scenes & Lines & Day 
            &24.72&0.550&0.528
            & 25.24 & 0.576 &0.496
			& 20.15  & 0.434  &0.717 
			&  20.29 & 0.434  & 0.711  \\
            26 & Buildings & Irregular & Day 
            &8.56 &0.242 &0.564 
            & 8.56 & 0.242 &0.564
			& 9.19  & 0.336  &0.924 
			& 9.23  & 0.341  & 0.913  \\
            27 & Citys & Irregular & Day 
            &4.54&0.006&0.705
            & 4.91 & 0.249 &0.818
			& 16.19  &0.443   &0.699 
			&  16.07 & 0.443  & 0.687  \\
            28 & Small area & Circles & Day 
            &24.48 &0.660 &0.479 
            & 24.32 &  0.630&0.493
			& 20.12  & 0.536  &0.643 
			&  21.13 & 0.595  & 0.621  \\
            29 & Buildings & Circle & Day 
            &22.98 &0.608 &0.540 
            & 23.58 & 0.631 &0.516
			& 16.57  & 0.439  &0.733 
			& 17.93  &  0.453 & 0.716  \\
            30 & Natural scenes & Irregular & Day 
            &20.23 &0.522 &0.605 
            & 21.02 & 0.559 &0.569
			& 12.36  & 0.431  &0.760 
			&15.40 & 0.450  &  0.719 \\
            31 & Citys & Circles & Night 
            &18.97&0.365&0.645
            & 19.09 & 0.371 &0.634
			& 17.88  & 0.465  &0.685 
			&  17.57 & 0.459  &0.704  \\
            32 & Citys & Irregular & Day 
            &17.99&0.582&0.621
            & 18.00 & 0.582 &0.628
			& 17.01  &0.623   &0.588 
			& 16.94  & 0.622  & 0.590  \\
            33 & Citys & Irregular & Day 
            &5.79 &0.007 &0.745 
            & 5.79 & 0.007 &0.744
			& 15.20  & 0.436  &0.761 
			& 14.68  &  0.431 &0.770   \\

            \midrule
            \textbf{Mean} &-&-&-
            &\textbf{18.72} &\textbf{0.484} &\textbf{0.600}
            & \textbf{19.01}  & \textbf{0.500}   &\textbf{0.592} 
			& \textbf{16.80}  &\textbf{0.489}   &\textbf{0.681} 
			&  \textbf{16.97} &\textbf{ 0.490}  &  \textbf{0.678} \\
			\bottomrule
		\end{tabular}
		}
		\vspace{-15pt}
	\end{center}
\end{table*}

\begin{table*}[t]
	\begin{center}
		\centering
		\caption{More sub-benchmarks for multi-modal NeRF synthesis. We divide our dataset into subsets based on different scene types, camera trajectories, and lighting conditions, and provide sub-benchmarks under different settings. $\uparrow$ means the higher, the better.}
		\label{tab:suppmultim2}
		\resizebox{1\textwidth}{!}{
		\begin{tabular}{cccccccccccccc}
			\toprule
			\multirow{2}{*}{Scene ID}  & \multirow{2}{*}{Sub-benchmark} & \multicolumn{3}{c}{\textbf{NeRF}~\cite{mildenhall2020nerf} w/o Prompts}& 
			\multicolumn{3}{c}{\textbf{NeRF}~\cite{mildenhall2020nerf} w/ Prompts}& 
			\multicolumn{3}{c}{\textbf{CoCo-INR}~\cite{yin2022coordinates} w/o Prompts}&
			\multicolumn{3}{c}{\textbf{CoCo-INR}~\cite{yin2022coordinates} w/ Prompts}\\
			 &  & PSNR$\uparrow$&SSIM$\uparrow$&LPIPS$\downarrow$& PSNR$\uparrow$&SSIM$\uparrow$&LPIPS$\downarrow$& 
			 PSNR$\uparrow$&SSIM$\uparrow$&LPIPS$\downarrow$& PSNR$\uparrow$&SSIM$\uparrow$&LPIPS$\downarrow$\\

		\cmidrule(l){1-1}  \cmidrule(l){2-2} \cmidrule(l){3-5} \cmidrule(l){6-8} \cmidrule(l){9-11} \cmidrule(l){12-14}

            1,4,7,8,11,13,22,26& \textbf{Buildings} 
            &17.32 & 0.501 & 0.605 		
            & 18.04  & 0.516  & 0.572 
            & 15.88  & 0.546  & 0.689 
            &  15.87 &  0.540 & 0.684  \\
            
            2,5,12,14,15,18,19,28,29& \textbf{Small areas} 
            & 20.88 & 0.579 & 0.561
            & 21.08  & 0.588  & 0.554 
            & 17.03  &0.501   & 0.665 
            &  17.56 & 0.515  & 0.656  \\
            
            3,8,9,10,20,27,31,32,33& \textbf{Cities} 
            &15.87 & 0.360 &0.637 		
            & 16.06  &  0.392 & 0.645 
            &  16.62 & 0.464  & 0.687 
            &16.54   & 0.464  & 0.687  \\
            
            6,16,17,21,23,24,25,30& \textbf{Natural scenes} 
            &21.23 	&0.503 	&0.597 		
            & 21.30  & 0.507  & 0.591 
            &  17.67 &0.437   & 0.686 
            & 17.84  & 0.431  &0.691   \\
            
            \midrule
            
            2,4,5,6,8,11,12,15,16,17,19,20,21,28,31& \textbf{Circles} 
            &21.08 	&0.573 	&0.559 		
            & 21.13  & 0.575  & 0.555 
            &  17.89 & 0.529  & 0.635 
            & 18.02  & 0.532  &  0.634 \\
            
            3,7,9,13,14,18,22,23,24,25& \textbf{Lines} 
            &18.24 	&0.450 	&0.628 		
            & 18.93  & 0.468  & 0.600 
            & 16.63  & 0.452  & 0.700 
            & 16.60  & 0.449  & 0.700  \\
            
            1,10,26,27,29,30,32,33& \textbf{Irregular} 
            &14.91 	&0.361 	&0.644 		
            & 15.15  & 0.398  & 0.649 
            & 14.96  & 0.460  & 0.743 
            & 15.46  & 0.463  &  0.734 \\
            
            \midrule

            ALL-\{4, 11,31\} & \textbf{Day} 
            &18.37 	&0.465 	&0.610 		
            & 18.69  & 0.482  & 0.602 
            & 16.59  & 0.468 & 0.694 
            & 16.77  & 0.469  & 0.690  \\
            
            4, 11,31& \textbf{Night} 
            &22.24 	&0.680 	&0.499 		
            & 22.20  & 0.681  & 0.490 
            & 18.86  &0.696   & 0.553 
            & 18.93  & 0.695  & 0.559  \\

			\bottomrule
		\end{tabular}
		}
	    \vspace{-9pt}
	\end{center}
\end{table*}

\clearpage

\section{Method Details}
\label{supp:method}

\begin{figure}[h]
	\centering
	\includegraphics[width=1\linewidth]{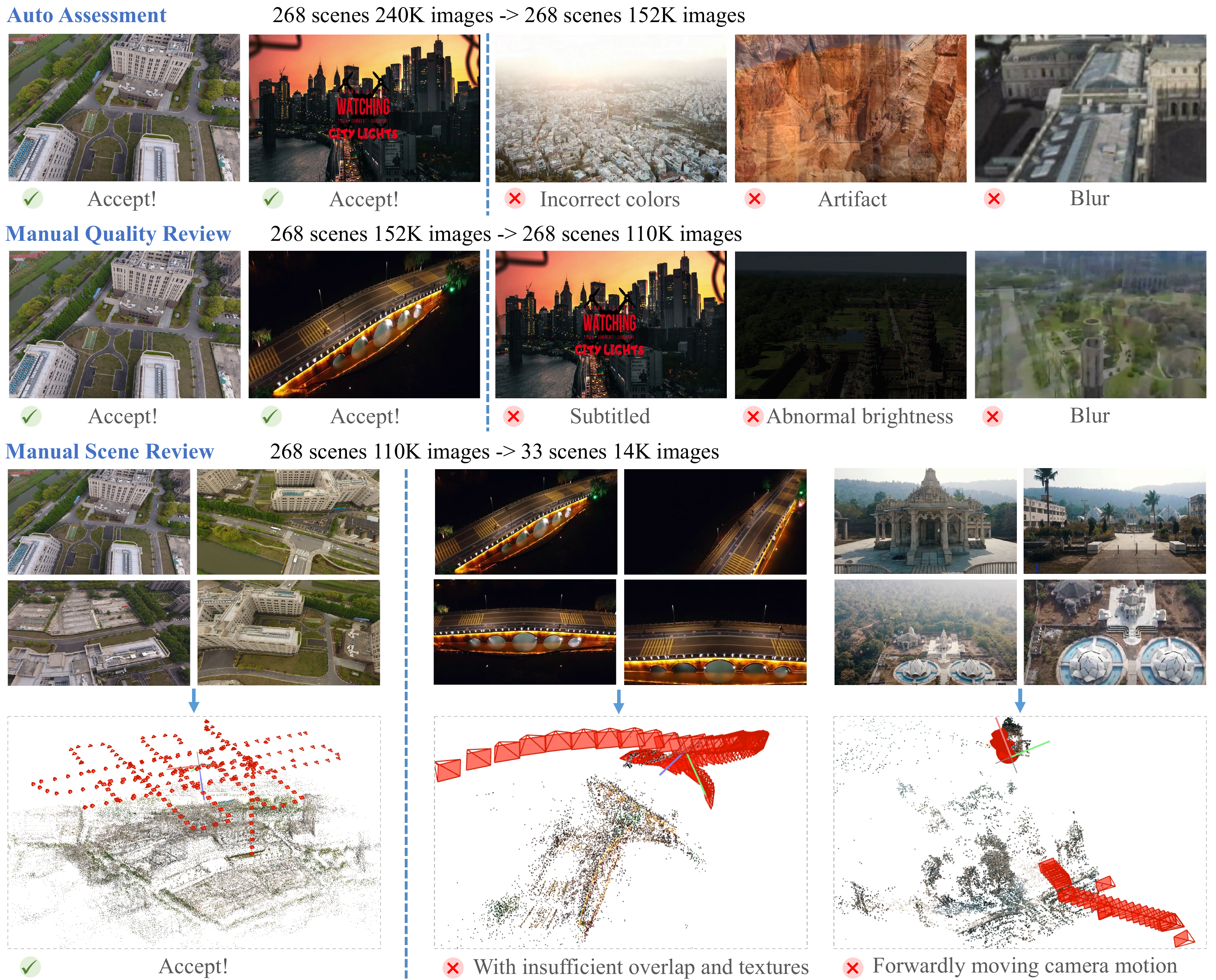}
	\caption{Some examples of dropped frames or scenes at auto assessment, manual quality review, and manual scene review stages. Meanwhile, we also show the number of images and scenes before and after the review at each stage.}
	\label{fig:suppbadcase}
\end{figure}

We show some dropped frames or scenes during dataset generation to better understand our selection and review standard in Fig.~\ref{fig:suppbadcase}. 
At auto assessment stage, the image quality assessment model~\cite{talebi2018nima} is employed to remove frames with blur, artifacts, ghosting and incorrect colors caused by overexposure or optical effects. In this way, about 64\% of the frames remained, but there are still some low-quality frames with blur, subtitles, abnormal brightness or transparency caused by fading in or out at the beginning or end of the video. So during the manual quality review process, volunteers and experts will work together to remove these frames. After scene calibration and reconstruction, some scenes will fail, such as with insufficient overlap and textures, or forwardly moving camera motion. These fail-to-calibrate scenes cannot meet the requirements of NeRF-based methods, which need to be removed at the manual scene review stage.

\clearpage

\section{Dataset Anaysis}
\label{supp:datasetAnaysis}

\subsection{Textual Prompts}

\begin{figure}[h]
	\centering
	\vspace{-3mm}
	\includegraphics[width=0.95\linewidth]{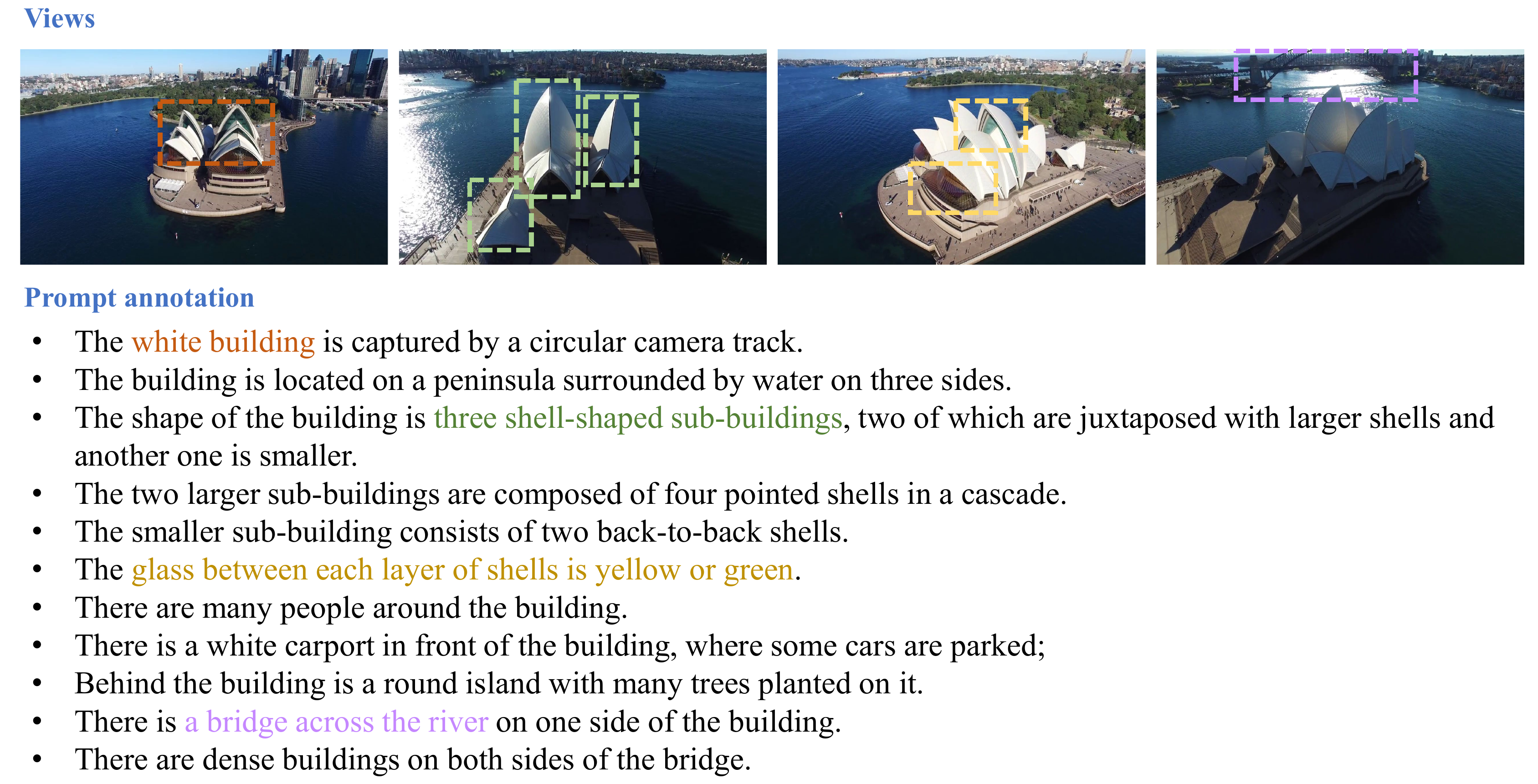}
	\vspace{-2mm}
	\caption{\textbf{Textual Prompt.} An example of annotations. Several phrases and their corresponding patches are highlighted in the same color.}
	\label{fig:supptext}
\end{figure}

We show an example of scene prompt annotations from our OMMO dataset in Fig.~\ref{fig:supptext}. Our prompts annotation comprehensively describes every detail of the scene center and its surrounding environment in many short sentences.

\begin{figure}[b]
	\centering
	\vspace{-3mm}
	\includegraphics[width=0.58\linewidth]{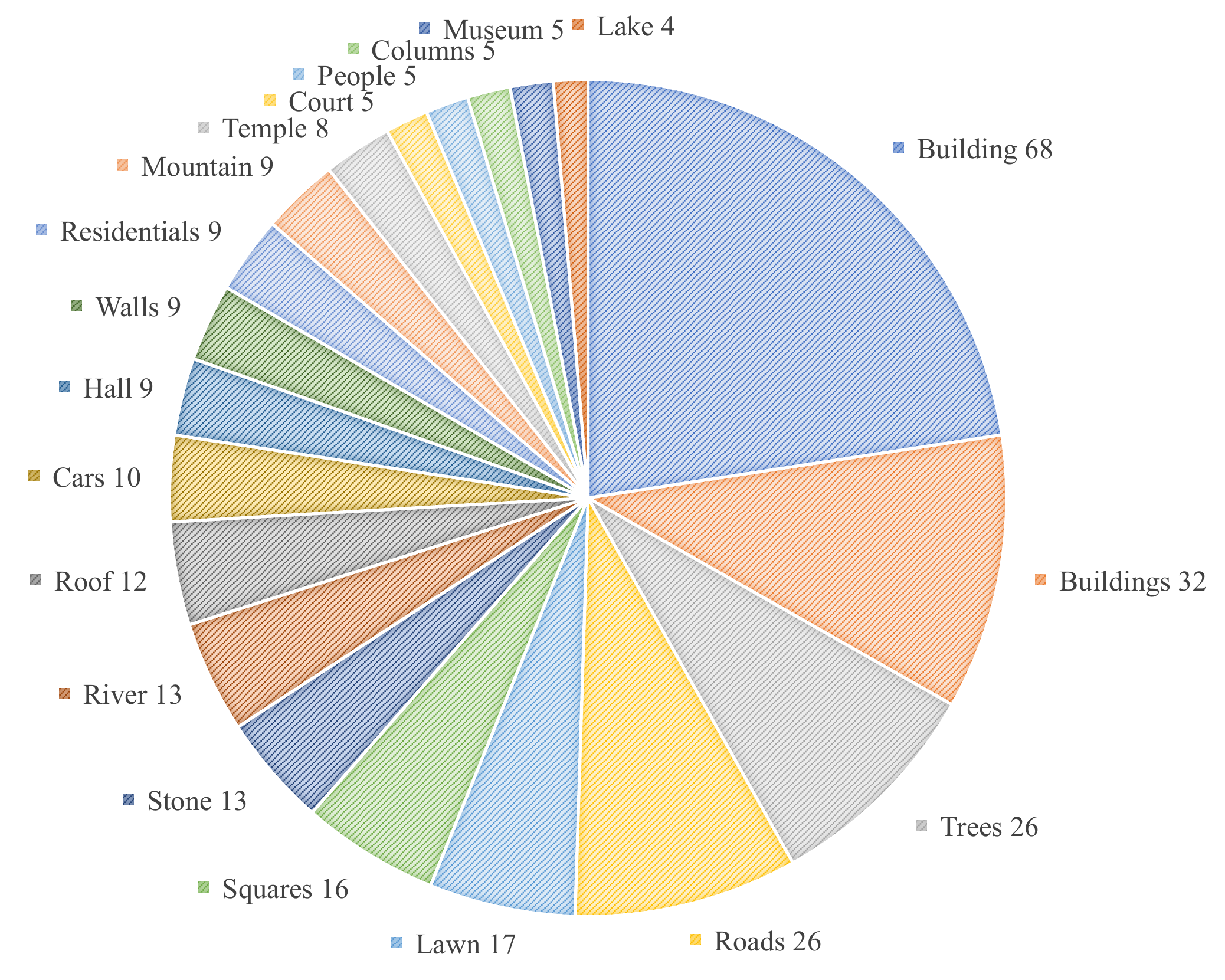}
	\vspace{-2mm}
	\caption{\textbf{Word statistic.} Only include nouns that appear more than 4 times in our OMMO dataset.}
	\label{fig:suppword}
\end{figure}

We report the word statistic for all scene prompts annotations (only including nouns that appear more than 4 times), as shown in Fig.~\ref{fig:suppword}. It can be seen that our data distribution is comprehensive and reasonable, including building, buildings (architectural complex), trees, roads, lawn and rivers, etc. Meanwhile, the number of keywords can roughly reflect the distribution of different scenes, such as natural scene: urban scene (building, small area, city) is about 1:3.

\clearpage
\subsection{User Instrictions}

    Our OMMO dataset structure list is shown below. The first-level directory contains the scene list, the training and validation split file, and sub-folders for each scene. Each scene contains original video and sub-folders for images, camera matrices and textual prompts.

\begin{figure}[h]
	\flushleft
	\vspace{-2mm}
	\includegraphics[width=0.25\linewidth]{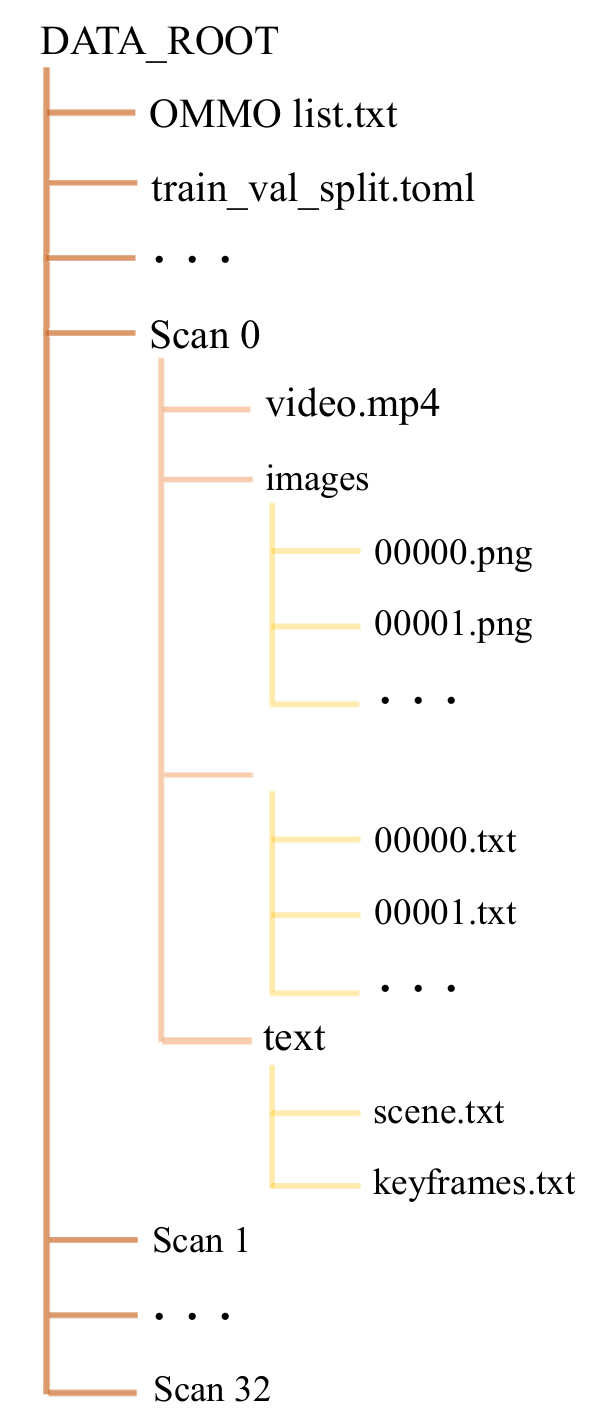}
\end{figure}

\end{document}